\documentclass[journal]{IEEEtran}

\usepackage{cite}
\usepackage{graphicx}
\usepackage[caption=false,font=footnotesize]{subfig}
\usepackage{amsmath}
\usepackage{amsfonts}
\usepackage{mathrsfs}
\usepackage{multirow}
\usepackage{booktabs}
\usepackage{algorithm}
\usepackage{algpseudocode}

\newcommand{\eg}{\emph{e.g.,}~}
\newcommand{\ie}{\emph{i.e.,}~}
\newcommand{\etal}{\emph{et al.}~}
\newcommand{\wrt}{\emph{w.r.t.}~}

\usepackage{color}

\begin{document}

\title{Fine-Grained Fashion Similarity Prediction by Attribute-Specific Embedding Learning}

\author{Jianfeng~Dong,
        Zhe~Ma,
        Xiaofeng~Mao,
        Xun~Yang,
        Yuan~He,
        Richang~Hong,
        and Shouling~Ji
\thanks{J. Dong is with the College of Computer and Information Engineering, Zhejiang Gongshang University, Hangzhou 310035, China. E-mail: dongjf24@gmail.com}
\thanks{Z. Ma is with the College of Computer Science and Technology, Zhejiang University, Hangzhou 310027, China. E-mail: maryeon.rs@zju.edu.cn}
\thanks{S. Ji is with the College of Computer Science and Technology, Zhejiang University, Hangzhou 310027, and Binjiang Institute of Zhejiang University, Hangzhou 310053, China. E-mail: sji@zju.edu.cn}
\thanks{X. Mao and Y. He are with the Alibaba Group, Hangzhou 311121, China. E-mail: mxf164419@alibaba-inc.com,heyuan.hy@alibaba-inc.com}
\thanks{X. Yang is with the School of Computing, National University of Singapore, Singapore 37580, Singapore. E-mail: xunyang@nus.edu.sg}
\thanks{R. Hong is with the School of Computer Science and
Information Engineering, Hefei University of Technology, Hefei 230009, China,
Email: hongrc.hfut@gmail.com}
\thanks{Jianfeng Dong and Zhe Ma are the co-first authors. Shouling Ji is the corresponding author. Manuscript received March 31, 2021; revised August 19, 2021; accepted September 16, 2021.}}

\markboth{IEEE Transactions on Image Processing, March~2021}{Shell \MakeLowercase{\textit{et al.}}: Bare Demo of IEEEtran.cls for IEEE Journals}

\maketitle

\begin{abstract}
This paper strives to predict fine-grained fashion similarity. In this similarity paradigm, one should pay more attention to the similarity in terms of a specific design/attribute between fashion items. For example, whether the collar designs of the two clothes are similar. It has potential value in many fashion related applications, such as fashion copyright protection. To this end, we propose an \textit{Attribute-Specific Embedding Network (ASEN)} to jointly learn multiple attribute-specific embeddings, thus measure the fine-grained similarity in the corresponding space.
The proposed ASEN is comprised of a global branch and a local branch. The global branch takes the whole image as input to extract features from a global perspective, while the local branch takes as input the zoomed-in region-of-interest (RoI) \wrt the specified attribute thus able to extract more fine-grained features. 
As the global branch and the local branch extract the features from different perspectives, they are complementary to each other.
Additionally, in each branch, two attention modules, \ie \textit{Attribute-aware Spatial Attention} and \textit{Attribute-aware Channel Attention}, are integrated to make ASEN be able to locate the related regions and capture the essential patterns under the guidance of the specified attribute, thus make the learned attribute-specific embeddings better reflect the fine-grained similarity. Extensive experiments on three fashion-related datasets, \ie FashionAI, DARN, and DeepFashion, show the effectiveness of ASEN for fine-grained fashion similarity prediction and its potential for fashion reranking. 
Code and data are available at https://github.com/maryeon/asenpp.
\end{abstract}

\begin{IEEEkeywords}
Fashion Retrieval, Fine-Grained Similarity, Fashion Understanding, Image Retrieval
\end{IEEEkeywords}

\section{Introduction}
 
\begin{figure}[!t]
\centering
\includegraphics[width=0.9\columnwidth]{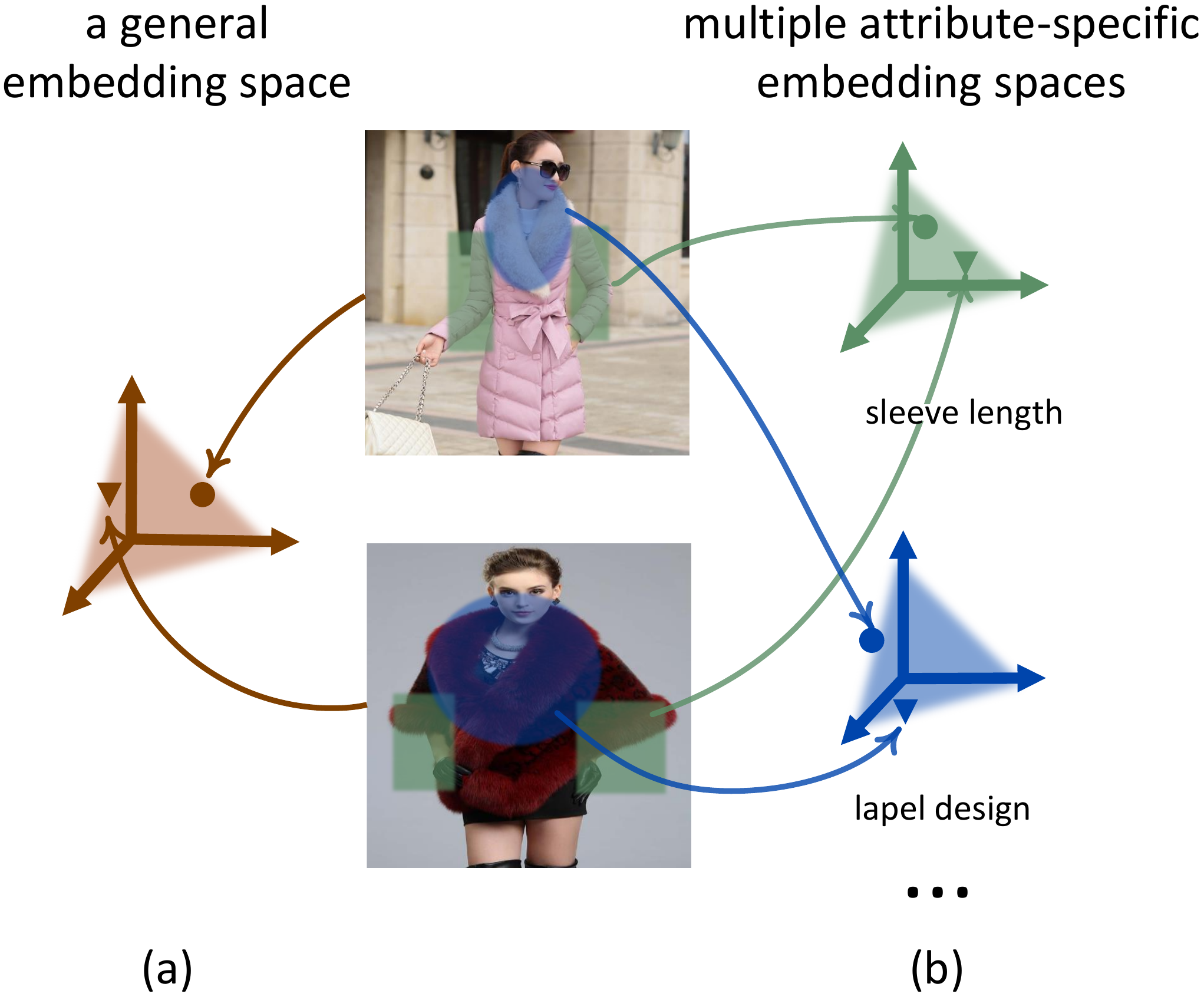}
\vspace{-3mm}
\caption{As fashion items typically have various attributes, we propose to learn multiple attribute-specific embeddings, thus the fine-grained similarity can be better reflected in the corresponding attribute-specific space.}
\vspace{-3mm}
\label{fig:concept}
\end{figure}

\IEEEPARstart{P}{redicting} the similarity between fashion items is essential for a number of fashion-related tasks including in-shop clothes retrieval \cite{liu2016deepfashion,ak2018efficient,wang2019multi,yang2019interpretable,ma2020knowledge}, street-to-shop cross-domain clothes retrieval \cite{liu2012street,huang2015DARNdataset,ji2017cross}, fashion compatibility prediction  \cite{he2016learning,vasileva2018learning,li2017mining,yang2019transnfcm,yang2020learning}, interactive fashion search\cite{zhao2017memory,ak2018learning,guo2018dialog,chen2020image,liao2018knowledge,yang2019interpretable} and so on.
In order to predict the fashion similarity, the majority of methods~\cite{zhao2017memory,ji2017cross,han2017learning} are proposed to learn a general embedding space where relevant pairs are forced to be closer than irrelevant ones, thus the similarity can be computed in the learned space by standard similarity metric, \eg cosine similarity.
As the above tasks aim to search for identical or similar/compatible fashion items with the query item, methods for these tasks tend to focus on the overall similarity~\cite{liu2016deepfashion,ji2017cross,chen2020image}.
In this paper, we aim for the fine-grained fashion similarity. Consider the two fashion images in Fig. \ref{fig:concept}, although they appear to be irrelevant overall, they actually present similar characteristics over some attributes, \eg both of them have the similar lapel design. We consider such similarity in terms of a specific attribute as the fine-grained similarity.

There are cases where one would like to search for fashion items with certain similar designs instead of identical or overall similar items, so the fine-grained similarity matters in such cases.
In the fashion copyright protection scenario~\cite{martin2019fashion}, the fine-grained similarity is also important to find items with plagiarized designs, as plagiarized items typically plagiarize parts of the original one~\cite{lang2020plagiarism}. Hence, learning the fine-grained similarity in the context of fashion is necessary.
However, to the best of our knowledge, such a similarity paradigm has been ignored by the community to some extent, few works~\cite{veit2017,tan2019learning} have focused on it. 
In~\cite{veit2017}, Veit \etal first learn an overall embedding space, and then employ fixed masks to select relevant embedding dimensions with respect to the specified attribute. The fine-grained similarity is measured in terms of the masked embedding features. As a follow-up work, Tan \etal~\cite{tan2019learning} propose to utilize dynamic weights to modulate the fixed masks.
In our work, we go further in this direction. 
As shown in \figurename \ref{fig:concept}(b), we propose to learn multiple attribute-specific embedding spaces where each corresponds to a specific attribute. Therefore, the fine-grained fashion similarity \wrt a specific attribute can be readily measured in the corresponding space. 
For example, from the perspective of \textit{lapel design}, the similarity between two clothes can be measured in the embedding space of \textit{lapel design}.

To this end, we propose an \textit{Attribute-Specific Embedding Network (ASEN)} to jointly learn multiple attribute-specific embeddings in an end-to-end manner.
\figurename \ref{fig_framework} illustrates the structure of our proposed ASEN, which is comprised of two branches \ie a global branch and a local branch.
The global branch takes the whole image as the input. Considering an attribute is typically related to the specific regions of the image, in the global branch we introduce a novel \textit{attribute-aware spatial attention (ASA)} which mainly focuses on the related regions with the specified attribute. Besides, the same regions may still be related to multiple attributes. For example, attributes \textit{collar design} and \textit{collar color} are all associated with the region around \textit{collar}. Therefore, we propose an \textit{attribute-aware channel attention (ACA)} module to extract important patterns. With both ASA and ACA, we expect the global branch being able to locate the related regions and capture the essential patterns \wrt the specified attribute.

Although the above attention mechanisms usually work well, they may still fail when the related regions are very small. We attribute it to the fact that the deep learning based networks typically transform input images into a fixed resolution despite the higher resolution of the original images. This process is able to improve the computational efficiency of models, while leads to losing some detailed information, especially for small objects.
To mitigate it, a local branch is also integrated into our proposed ASEN. Instead of the whole image, the local branch takes a zoomed-in region-of-interest (RoI) \wrt the specified attribute of the given image as the input. This allows the local branch for extracting more fine-grained features.
The RoI is obtained by a weakly-supervised localization method, which resorts to the attribute-aware spatial attention of the global branch and regards the highly activated region of the attention map as the RoI. The weakly-supervised localization method does not need any supervised bounding box annotations, is easy and effective.
Note that the local branch shares a similar network architecture with the global branch, but with different parameters to fit the inputs of different resolutions.

The features from both global and local branches are concatenated as the attribute-specific features, and the final fine-grained fashion similarity is measured in terms of the attribute-specific features.
It is worth pointing out that fine-grained similarity learning is orthogonal to overall similarity learning, allowing us to utilize ASEN to facilitate traditional fashion retrieval that focuses on the overall similarity, such as in-shop clothes retrieval.

In sum, this paper makes the following contributions:
\begin{itemize}
\item Conceptually, we propose to learn multiple attribute-specific embedding spaces for fine-grained fashion similarity prediction. As such, a certain fine-grained similarity between fashion items can be measured in the corresponding space. 
\item Technically, we propose a novel ASEN model consisting of a global branch and a local branch to effectively realize the above proposal. Combined with ASA and ACA, both branches extract essential features under the guidance of the specified attribute, which benefits the fine-grained similarity computation.
\item Based on the proposed attention mechanism, we devise a weakly-supervised localization method to localize RoI \wrt the specified attribute. Besides, we also design a two-stage training strategy to train two-branch ASEN.
\item Extensive experiments on FashionAI \cite{zou2019fashionai}, DARN \cite{huang2015DARNdataset} and DeepFashion~\cite{liu2016deepfashion} datasets demonstrate the effectiveness of ASEN for fine-grained fashion similarity prediction and its potential for fashion reranking.
\end{itemize}

Different from the majority of works that tend to learn a general embedding space for the overall similarity prediction, the main novelty of this work is proposing to learn multiple attribute-specific embedding spaces for fine-grained fashion similarity prediction. Additionally, to realize the above proposal, we devise a two-branch model architecture ASEN with the components of two attribute-aware attentions, a weakly-supervised localization method and a two-stage training strategy. Our another novelty is the integration of these components into a two-branch manner for fine-grained fashion similarity prediction. We consider the novelty of the system architecture to be more than the sum of its parts. The ASEN network effectively combines these components into a powerful solution that is the new state-of-the-art on three datasets for fine-grained fashion retrieval.

A preliminary version of this work was published at AAAI 2020~\cite{ma2020fine}. The journal extension mainly improves over the conference paper in the following aspects. 
Technically, we extend the single-branch ASEN to a new ASEN model consisting of a global branch and a local branch. A weakly-supervised localization method and a two-stage training strategy are also devised to fulfil the two-branch model.
Experimentally, besides the mean average precision, an extra performance metric Recall@100 is employed to evaluate models.
Compared to the ASEN with only one branch proposed in \cite{ma2020fine}, the new model with two branches consistently leads to better fine-grained fashion retrieval performance on three datasets. Additionally, more extensive experiments are included: exploring the viability of each component, the influence of the hyper-parameters in ASEN, the complexity analysis of the proposed model.
Based on the above improvement, we extend the conference version of 8 pages to the journal version of 15 pages.

\section{Related Work}\label{sec:relwork}

\subsection{Fashion Similarity Prediction}
In the following, we review recent progress on fashion similarity in the context of in-shop clothes retrieval, street-to-shop clothes retrieval, and fashion compatibility prediction.

Given a query fashion image, in-shop clothes retrieval task aims to retrieve images containing identical or similar clothes from a gallery of candidate images. As all the images are taken in shops and from the same domain, metric learning based solution are dominated to predict the fashion similarity~\cite{wang2017clothing,shankar2017deep,kinli2019fashion,wang2019multi}. For instance, Kinli \etal \cite{kinli2019fashion} devised a capsule network based model to extract features for both query and candidate images, and a triplet ranking loss was employed to train the model. Instead of the triplet ranking loss, Wang \etal \cite{wang2019multi} proposed a multi-similarity loss which jointly considers self similarity and relative similarity of training samples.

In contrast to in-shop clothes retrieval where images are from the same domain, images in street-to-shop clothes retrieval are from two different domains thus the task is more challenging.
In an early work~\cite{liu2012street}, Liu \etal first decomposed fashion images into multiple human parts, and images from the street domain are represented by collaborative sparse reconstruction based on the online shop images and an auxiliary fashion image set. As it utilizes hand-crafted human part features, its performance is suboptimal.
Recently,  with the success of deep learning in computer vision, we observe that recent efforts on street-to-shop fashion retrieval are typically based on deep learning techniques.
A popular pipeline is to learn a street-shop common space, and the similarity between images from two domains can be measured in the learned space~\cite{chen2015DDANdataset,huang2015DARNdataset,kiapour2015WTBIdataset,shankar2017deep,wang2017clothing,Gajic2018CrossDomainFI}.
For instance, Huang \etal \cite{huang2015DARNdataset} respectively utilized two Convolutional Neural Networks (CNN) based branches for two domains, and projected them into a common embedding space. 
Kuang \etal \cite{kuang2019fashion} used a Graph Reasoning Network to build a similarity pyramid which considers both global and local representations of fashion images.
Different from \cite{huang2015DARNdataset,kuang2019fashion} that extract features solely from images, Liu \etal~\cite{liu2016deepfashion} resorted to the fashion landmark prediction, Ji \etal \cite{ji2017cross} additionally employed semantic tags associated with images. More specifically, in \cite{liu2016deepfashion}, Liu \etal utilized fashion landmark prediction as an auxiliary task, and the predicted landmarks are used to spatially pool or gate the learned features to increase the discrimination of features.
Ji \etal \cite{ji2017cross} exploited the rich semantic tags as the guidance to focus on the important region of fashion images.

In the context of fashion compatibility prediction task, the fashion similarity is modeled as the visually compatibility or functionally complementarity between fashion items, typically between different categories~\cite{liu2012hi,veit2015learning,mcauley2015image,he2016learning,tangseng2017recommending,li2017mining,han2017learning,vasileva2018learning,lin2020fashion}.
For instance, in \cite{tangseng2017recommending}, Tangseng \etal formulated the fashion prediction task as a binary classification problem, and proposed a CNN based classification model. The proposed model directly takes fashion items as the input, and predicts whether the fashion items are compatible.
Similarly, Li \etal \cite{li2017mining} also proposed a binary classification model, but additionally utilized meta-data of fashion items, such as title and category, to further enhance fashion representations. 
Different from \cite{tangseng2017recommending,li2017mining} that are classification based solutions, Veit \etal \cite{veit2015learning} utilized a neural network to map fashion items into an embedding space, thus predict whether two input fashion items are compatible according to their distance in the space. 
Considering two fashion items that are asked to predict compatibility are typically in different categories, Vasileva \etal \cite{vasileva2018learning} proposed to learn multiple embedding spaces corresponding to different category pairs instead of one embedding space. 
With the similar idea of \cite{vasileva2018learning}, Lin \etal \cite{lin2020fashion} first learned multiple embedding spaces, but additionally utilized a category-guided attention to aggregate multiple embedding features.
Note that although our proposed model also learn multiple embedding spaces, our learned embedding spaces correspond to attributes and are used for fine-grained fashion similarity measurement. 

Different from the above methods that focus on the overall similarity (identical or overall similar/compatible), we study the fine-grained similarity in the paper. In \cite{veit2017}, Veit \etal have made the first attempt in this direction. In their approach, an overall embedding space is first learned, and the fine-grained similarity is measured in this space with the fixed mask w.r.t. a specified attribute. 
By contrast, we jointly learn multiple attribute-specific embedding spaces, and measure the fine-grained similarity in the corresponding attribute-specific space. It is worth noting that \cite{vasileva2018learning,he2016learning} also learned multiple embedding spaces, but they still focus on the overall similarity.

\subsection{Attention Mechanism}
Recently, attention mechanism has become a popular technique and showed superior effectiveness in various research areas, such as computer vision \cite{woo2018cbam,wang2017residual,qiao2018exploring,wu2019hierarchical,wang2018non,bello2019attention}, natural language processing \cite{bahdanau2014neural,vaswani2017attention,devlin2018bert} and multimedia \cite{liu2020jointly,he2018nais,qu2020fine,yin2019multi,liu2021context}.
To some extent, attention can be regarded as a tool to bias the allocation of the input information.
As fashion images always present with complex backgrounds, pose variations, deformations, \textit{etc.}, attention mechanism was also common in the fashion domain \cite{ji2017cross,wang2017clothing,han2017automatic,ak2018learning,ak2018efficient,liu2016deepfashion}. 
For instance, Ak\etal \cite{ak2018efficient} used the prior knowledge of clothes structure to locate the specific parts of clothes.
However, their approach can be only used for upper-body clothes thus limits its generalization. In \cite{wang2017clothing}, Wang \etal proposed to learn a channel attention implemented by a fully convolutional network.
Instead of directly learn the attention map, Ak \etal \cite{ak2018learning} generated attention map by calculating activation maps based on attribute classification results~\cite{zhou2016learning}.
In \cite{han2017automatic}, Han \etal utilized the similar attention generation for concept discovery in the fashion domain.
The above attentions are in a self-attention manner without explicit guidance for attention mechanism.
In this paper, we propose two attribute-aware attention modules, which utilize a specific attribute as the extra input in addition to a given image. The proposed attention modules capture the attribute-related patterns under the guidance of the specified attribute.
Note that Ji \etal \cite{ji2017cross} also utilized attributes to facilitate attention modeling,
but they use all attributes of fashion items and aim for learning a better discriminative fashion feature. 
By contrast, we employ each attribute individually to obtain more fine-grained attribute-aware feature for fine-grained similarity computation.

\subsection{Fine-grained Image Retrieval}
Our work is also related to fine-grained image analysis~\cite{wei2019deep}, especially for fine-grained image retrieval task~\cite{wei2017selective,zheng2018centralized,zheng2019towards,chen2020exploration,cui2020exchnet}.
Given a query image, this task aims to retrieve images of the same category as the query image, from fine-grained candidate images that contain only subtle differences (\eg varieties of birds).
For instance, as the first work of fine-grained image retrieval using deep learning techniques, Wei \etal~\cite{wei2017selective} first selected the meaningful deep descriptors of fine-grained images by pre-trained CNN models, and then aggregated the selected descriptors to represent images.
In \cite{zheng2018centralized}, Zheng \etal proposed to extracts more discriminative object-level features by object localization, and a novel centralized ranking loss was employed.
As retrieval speed and storage cost matter for retrieval tasks, Cui \etal~\cite{cui2020exchnet} recently proposed a unified end-to-end trainable hashing network to represent images into compact binary codes.
Different from the above works that tend to extract object-level features, our work aims to extract more fine-grained features, for example, collar design of clothing.

\section{Proposed Method}

\begin{figure*}[!t]
\centering
\includegraphics[width=0.9\textwidth]{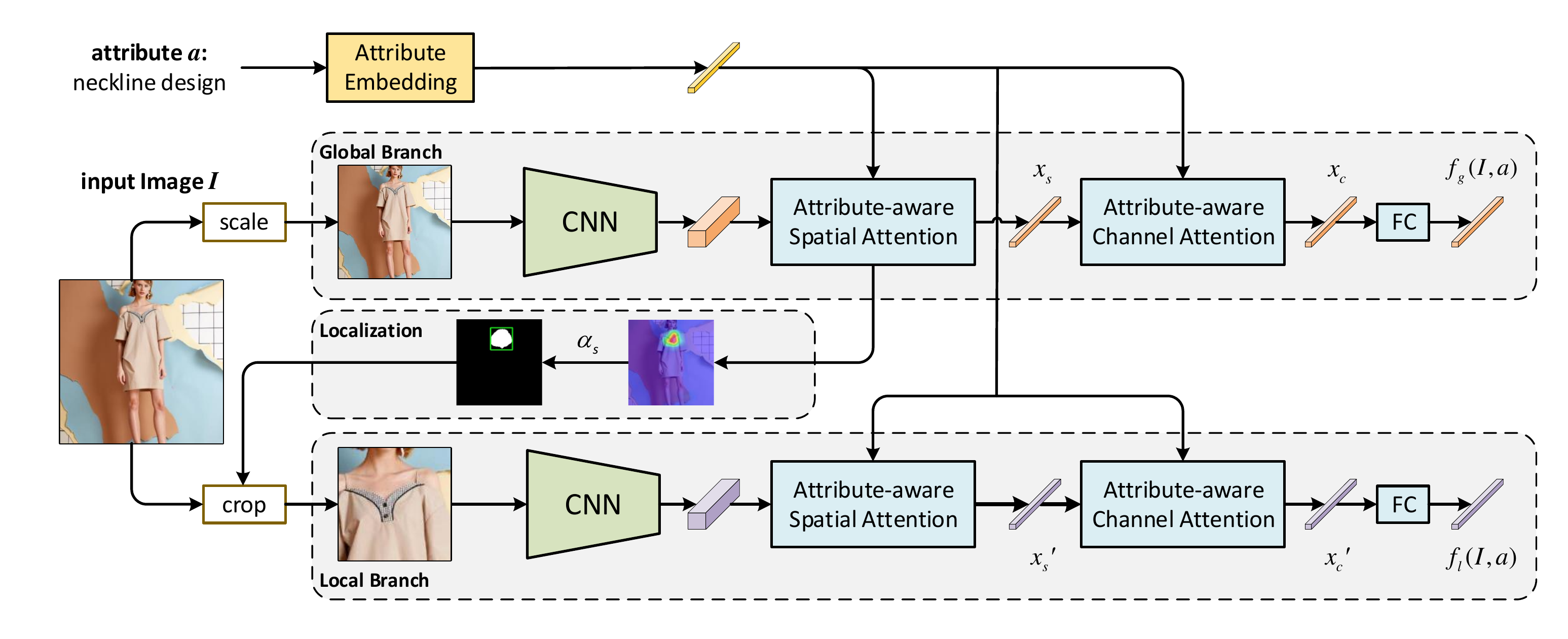}
\vspace{-3mm}
\caption{The framework of our proposed Attribute-Specific Embedding Network (ASEN) which consists of a global branch and a local branches. The global branch takes a whole image as input, while the local branch takes a RoI \wrt the specified attribute instead of the whole image as input.   The RoI is obtained by a weakly-supervised localization method,  which resorts to the attribute-aware  spatial attention of the global branch and regards the highly activated region of the attention map  as the RoI. 
Each branch is composed of a backbone network, an attribute-aware spatial attention and an attribute-aware channel attention modules. The whole network is trained by a two-stage training strategy.}
\vspace{-3mm}
\label{fig_framework}
\end{figure*}

In this section, we first give an overview of the proposed ASEN model consisting of a global branch and a local branch, then respectively elaborate the structure of the global branch and the local branch. The learning and inference of ASEN are introduced in Section \ref{ssec:learnig}.

\subsection{The Overview of ASEN}
Given an image $I$ and a specific attribute $a$, we propose to learn an attribute-specific feature vector $f(I,a)$ which reflects the characteristics of the given attribute in the image.
We extract the feature from both a global view and a local view, resulting in a global attribute-specific feature $f_g(I,a)$  and a local attribute-specific feature $f_l(I,a)$. Both the global and local features are combined as the final attribute-specific feature vector $f(I,a)$.
Therefore, given two fashion images, their fine-grained fashion similarity with respect to the attribute $a$ can be computed in terms of their corresponding attribute-specific feature vectors.

\figurename \ref{fig_framework} illustrates the framework of our proposed Attribute-Specific Embedding Network (ASEN) that is composed of two branches, a global branch $f_g$ and a local branch $f_l$. 
The global branch takes a whole image as input, while the local branch takes a RoI \wrt the specified attribute as input.
In each branch, it has a CNN based feature extraction module, followed by an attribute-aware spatial attention (ASA) and an attribute-aware channel attention (ACA) to capture spatial and channel dependencies with respect to the given attribute respectively. Both branches have the similar network architecture, but with different parameters to fit the inputs with different resolutions. 
As the global branch and the local branch extract the features from different perspectives, they are complementary to each other.
Additionally, the local branch is dependent on the global branch, which resorts to the attribute-aware spatial attention of the global branch and regards the highly activated region of the attention map as the RoI with the given attribute.

\subsection{Global Branch}
Given an image $I$ and a specific attribute $a$, the global branch learns a global attribute-specific feature vector $f_g(I, a)\in \mathbb{R}^{c_o}$ which reflects the characteristics of the corresponding attribute in the whole input image.
The global branch is composed of a feature extraction module, an ASA module and an ACA module. 
In what follows, we first detail the feature extraction, followed by the description of two attribute-aware attention modules.

\subsubsection{Feature Extraction}
To represent an input image $I$, 
we employ a CNN model pre-trained on ImageNet \cite{deng2009imagenet} as a backbone network, \ie ResNet-50 \cite{he2016deep}.
To keep the spatial information of the image, we remove the last fully connected layer in the CNN. Besides, we also discard convolutional block named as \textit{conv5}\_x to obtain relatively larger feature maps. Finally, the image is represented by $x \in \mathbb{R}^{c\times h\times w}$, where $h\times w$ is the size of the feature maps, $c$ indicates the number of channels.
For an attribute $a$, we first represent it by a one-hot vector, followed by a word embedding matrix $A \in \mathbb{R}^{n\times c_a}$, where $n$ denotes the total number of different attributes and $c_a$ is the dimensionality of attribute embedding. Finally, the attribute can be represented by a real-value vector $a \in \mathbb{R}^{c_a}$.

\subsubsection{Attribute-aware Spatial Attention (ASA)}
Considering the attribute-specific feature is typically related to the specific regions of the image, we only need to focus on the certain related regions. For instance, in order to extract the attribute-specific feature of the \textit{neckline design} attribute, the region around \textit{neck} is much more important than the others.
Besides, as fashion images always show up in large variations, \eg various poses and scales, using a fixed region with respect to a specific attribute for all images is not optimal.
Hence, we propose ASA which adaptively attends to certain regions of the input image under the guidance of the given attribute.
Given image representation $x$ and attribute representation $a$, we obtain the spatially attended vector \wrt the given attribute $a$ by $x_s = Attn_{s}(x,a)$, where the attended vector is computed as the weighted sum of input image feature vectors according to the given attribute.
Specifically, we first transform the image and the attribute to make their dimensionality same.
For the image, we employ a convolutional layer followed by a nonlinear $tanh$ activation function.
Formally, the mapped image $p(x)\in \mathbb{R}^{c_1\times h\times w}$ is given by 
\begin{equation}
p(x) = tanh(Conv_{c_1}(x)),
\end{equation}
where $Conv_{c_1}$ indicates a convolutional layer that contains $c_1$ $1\times 1$ convolution kernels.
For the attribute, we first project it into a $c_1$-dimensional vector through a non-linear mapping, implemented by a Fully Connected (FC) layer with a $tanh$ activation function, then perform spatial duplication. Hence, the mapped attribute $p(a) \in \mathbb{R}^{c_1 \times h\times w}$ is
\begin{equation}
p(a) = tanh(W_s a)\cdot \textbf{1},
\end{equation}
where $W_s\in \mathbb{R}^{c_1\times c_a}$ denotes the transformation matrix and $\textbf{1}\in \mathbb{R}^{1\times h\times w}$ indicates spatially duplicate matrix.
After the feature mapping, the attention weights $\alpha^s \in \mathbb{R}^{h\times w} $ is computed as:
\begin{equation}
\alpha^s = softmax(\frac{\sum_i^{c_1} [p(a) \odot p(x)]_i}{\sqrt{c_1}}),
\label{eq_attention_map}
\end{equation}
where $\odot$ indicates the Hadamard product, channel information is aggregated by a channel-wise sum. As suggested by \cite{vaswani2017attention}, we employ a scaling factor $\frac{1}{\sqrt{c_1}}$ in the attention. We also utilize a $softmax$ layer to normalize the attention weights.
With adaptive attention weights, the spatially attended feature vector of the image $I$ with respect to the given attribute $a$ is calculated as:
\begin{equation}\label{eq_spatial_attention}
x_s = \sum_{j}^{h\times w} \alpha^{s}_{j} x_j,
\end{equation}
where $\alpha^{s}_{j} \in \mathbb{R}$ and $x_j \in \mathbb{R}^{c}$ are the attention weight and the image feature vector at location $j$ of $\alpha^{s}$ and $x$ respectively.

Note that our ASA is inspired by attention mechanisms in vision and language related tasks~\cite{peng2019mava,nam2017dual}, such as VQA~\cite{xu2016ask,anderson2018bottom}, where a natural language sentence is typically employed as the guidance to locate related regions in images. In ASA, we utilize an attribute as the guidance to extract attribute-aware features, which is crucial for fine-grained similarity prediction.

\subsubsection{Attribute-aware Channel Attention (ACA)}
Although ASA adaptively focuses on the specific regions in the image, the same regions may still be related to multiple attributes. For example, attributes \textit{collar design} and \textit{collar color} are all associated with the region around \textit{collar}.
Hence, we further employ attribute-aware channel attention over the spatially attended feature vector $x_s$. 
ACA is designed as an element-wise gating function which selects the relevant dimensions of the spatially attended feature with respect to the given attribute.
Considering the different purposes of the two attentions, we use different attribute features in ASA and ACA.
Therefore, we additionally employ an attribute mapping layer to project attribute $a$ into a new $c_2$-dimensional feature vector, that is:
\begin{equation}
q(a) = \delta(W_c a),
\end{equation}
where $W_c\in \mathbb{R}^{c_2\times c_a}$ denotes the mapping matrix parameters and $\delta$ refers to \textit{ReLU} function. 
Then the mapped attribute feature and the spatially attended feature are fused by simple concatenation, and further fed into the subsequent two FC layers to obtain the attribute-aware channel attention weights.
As suggested in~\cite{hu2018squeeze}, we implement the two FC layers by a dimensionality-reduction layer with a reduction rate $r$ and a dimensionality-increasing layer, which has fewer parameters than one FC layer while introducing non-linear components. 
Formally, the attention weights $\alpha^c \in \mathbb{R}^c$ is calculated by:
\begin{equation}\label{eq_channel_attention}
    \alpha^c = \sigma(W_2 \delta(W_1 [q(a),x_s])),
\end{equation}
where $[,]$ denotes concatenation operation, $\sigma$ indicates \textit{sigmoid} function, $W_1\in \mathbb{R}^{\frac{c}{r}\times (c+c_2)}$ and $W_2\in \mathbb{R}^{c\times \frac{c}{r}}$ are transformation matrices. Here we omit the bias terms for description simplicity.
The final output of the ACA is obtained by scaling $x_s$ with the attention weight $\alpha^c$:
\begin{equation}
    x_c = x_s \odot \alpha^c.
\end{equation}

Finally, we further employ a FC layer over $x_c$ to generate the global attribute-specific feature of the given image $I$ with the specified attribute $a$:
\begin{equation}
f_g(I, a) = W x_c + b,
\end{equation}
where $W\in \mathbb{R}^{c_o\times c}$ is the transformation matrix, $b\in \mathbb{R}^{c_o}$ indicates the bias term, $c_o$ is the output dimensionality.

Although our ACA seems to be similar to the channel attention in SENet~\cite{hu2018squeeze}, we argue the main difference is as follows. With an image as the input, the channel attention in SENet is a kind of self-attention mechanism, which re-weights the channel-wise feature responses to extract salient or important information without any specific guidance. By contrast, besides the input image, ACA utilizes a specific attribute as the extra input. Therefore, our attention module is attribute-dependent, which is designed to capture the attribute-related patterns under the guidance of the specified attribute.

\subsection{Local Branch}
\begin{figure}[!t]
\centering
\vspace{-3mm}
\includegraphics[width=\columnwidth]{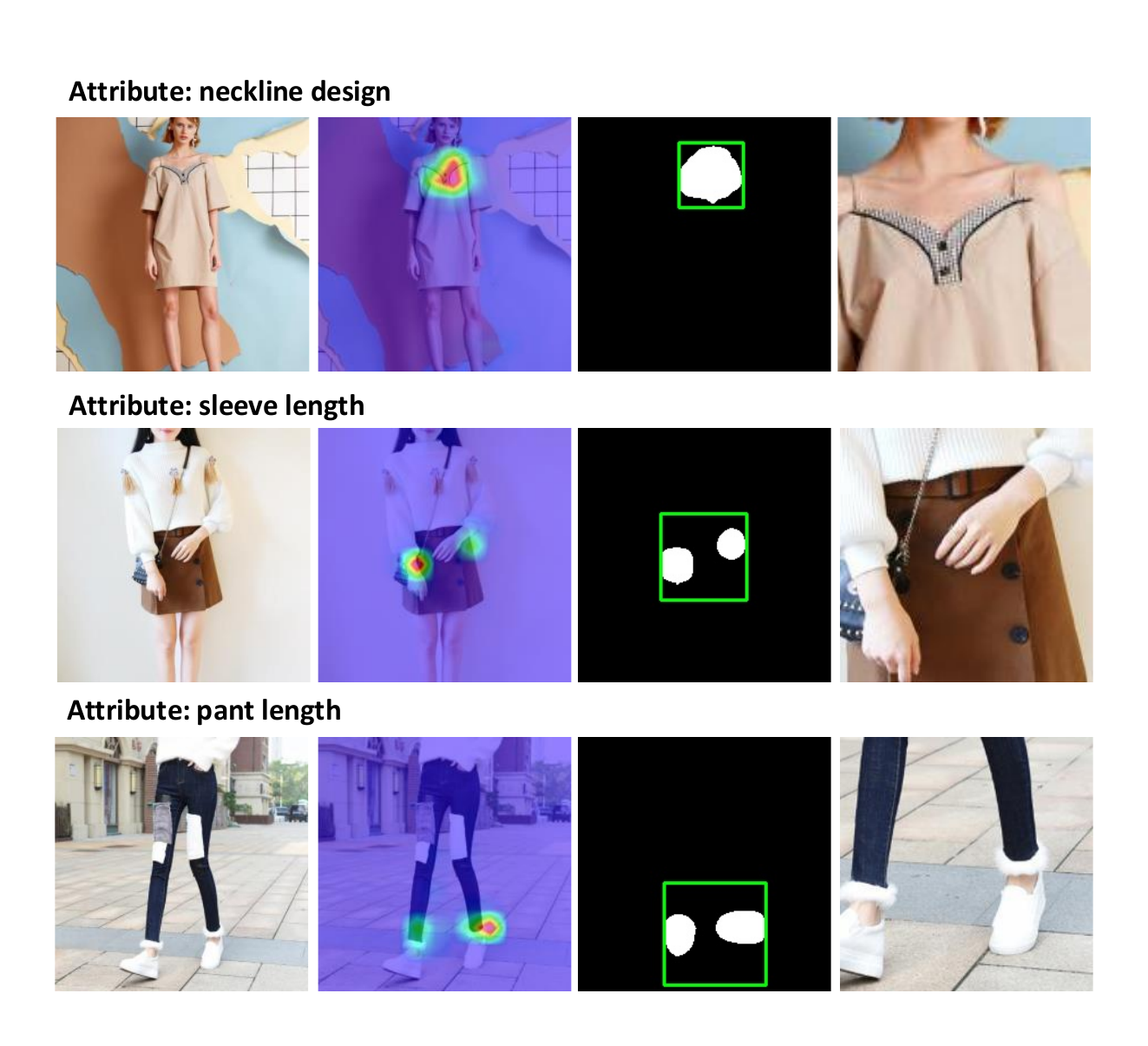}
\caption{Three examples of obtaining the ROI \wrt the given attribute by weakly-supervised localization method. For each example, four images are the original image, the attention map of ASA in the global branch, the binary map, and the cropped RoI, respectively.}
\vspace{-3mm}
\label{fig_crop}
\end{figure}

The structure of the local branch is similar with that of the above global branch, except for minor modifications on the input.
Given an image $I$ and an attribute $a$, the local branch predicts the local attribute-specific feature vector $f_l(I, a)\in \mathbb{R}^{c_o}$ based on a RoI of the input image instead of the whole input image. 
In order to obtain the RoI of the input image with the given attribute, we devise a weakly-supervised localization method that does not need any supervised bounding box annotations. The weakly-supervised localization method resorts to ASA module in the global branch, considering it typically locates the relevant regions with respect to the given attribute well.
To be Specific, given an attention map $\alpha^s$ generated by ASA module in the global branch, we first up-sample it to the same size of the input image $I$, followed by a binarization with a threshold to obtain a binary map. 
Since we feed a square RoI into the local branch, we first extract the minimal bounding box of non-zero pixels in the binary map, and further extend the shorter side of the bounding box to be square and make the non-zero pixels in the middle.
Finally, the region in the extracted square bounding box is regarded as the RoI of the input image.
\figurename \ref{fig_crop} illustrates some examples of obtaining the RoI.
Once the location of RoI is hypothesized, we crop and zoom in the RoI to a finer scale with higher resolution to obtain more fine-grained attribute-specific features. Similar to the global branch, the feature extraction, ACA and ASA modules are successively employed over the cropped RoI, resulting in $f_l(I, a)$. Note that the attribute embedding is shared for the global branch and the local branch.

\subsection{Learning and Inference}\label{ssec:learnig}
Based on the aforementioned framework, we then formulate the learning and inference process.

\subsubsection{Learning}
We would like to achieve multiple attribute-specific embedding spaces where the distance in a particular space is small for images with the same specific attribute value, but large for those with different ones. Consider the \textit{neckline design} attribute for instance, we expect the fashion images with \textit{Round Neck} near those with the same \textit{Round Neck} in the \textit{neckline design} embedding space, but far away from those with \textit{V Neck}. To this end, we choose to use the triplet ranking loss which is consistently found to be effective in various embedding learning related tasks \cite{yang2018person,dong2019dual,yang2017person,schroff2015facenet,yang2020tree,li2019w2vv++,liu2021hierarchical}. 
In what follows, we first detail the losses of our proposed model, followed by the description of our training strategy.

\textit{Losses.}
We employ the triplet ranking loss on both the global branch and the local branch. Specifically, we first construct a set of triplets $\mathcal{T} = \{(I, I^+, I^- | a)\}$, where $I^+$ and $I^-$ respectively indicate images relevant and irrelevant with respect to image $I$ in terms of attribute $a$.
For the global branch, the triplet ranking loss over a minibatch $ \mathcal{B} = \left\{\left( I, I^+, I^- | a \right)\right\}$ sampled from $\mathcal{T}$ is defined as:
\begin{equation}\label{eq_rank_loss_g}
\mathcal{L}_g = \sum_{\left( I, I^+, I^- | a \right) \in \mathcal{B} }  \max\left (0,m - s_g(I,I^+ | a)+s_g(I,I^- | a)\right ),
\end{equation}
where $m$ represents the margin, empirically set to be 0.2.
$s_g(I,I^* | a)$ denotes the global fine-grained similarity \wrt the attribute $a$, which can be expressed by the cosine similarity between $f_g(I, a)$ and $f_g(I^*,a)$, namely:
\begin{equation}
s_g(I,I^* | a)=\frac{f_g(I, a)\cdot f_g(I^*,a)}{\left\|f_g(I, a)\right\|_2\left\|f_g(I^*,a)\right\|_2}.
\end{equation}
Similarly, the triplet ranking loss of the local branch is computed as:
\begin{equation}\label{eq_rank_loss_l}
\mathcal{L}_l = \sum_{\left( I, I^+, I^- | a \right) \in \mathcal{B} } \max(0,m - s_l(I,I^+ | a)+s_l(I,I^- | a)),
\end{equation}
$s_l(I,I^* | a)$ denotes the local fine-grained similarity \wrt the attribute $a$, which is computed as the cosine similarity between the local attribute-specific feature vectors $f_l(I, a)$ and $f_l(I^*,a)$.

Furthermore, we also add an alignment loss between the global branch and the local branch, which encourages original images and the corresponding localized regions to be consistently embedded. It can be implemented by minimizing the difference between the global attribute-specific feature and the local attribute-specific feature. Given a triplet, we employ the loss over the three images. Formally, the alignment loss is defined as:
\begin{equation}
    \mathcal{L}_a =  \sum_{\left( I, I^+, I^- | a \right) \in \mathcal{B} } \left[ d_{gl}(I, a) + d_{gl}(I^+,a) + d_{gl}(I^-,a)\right],
\end{equation}
where $d_{gl}(I^*,a)$ denotes difference between the global attribute-specific feature and the local attribute-specific feature of image $I^*$ \wrt the attribute $a$, namely:
\begin{equation}
    d_{gl}(I^*,a) = 1 - \frac{f_g(I^*,a)\cdot f_l(I^*,a)}
{\left\|f_g(I^*,a)\right\|_2\left\|f_l(I^*,a)\right\|_2}.
\end{equation}

\textit{Training strategy.}
In order to train our proposed model, we perform the following two-stage training strategy: 
(1) We first train the global branch by minimizing the triplet loss defined in Eq. \ref{eq_rank_loss_g}. After that, the global branch is usually able to generate satisfactory RoI for the local branch.
(2) We train the whole model by minimizing the following joint loss:
\begin{equation}\label{eq_obj}
\mathcal{L} = \alpha \mathcal{L}_{g} + \beta \mathcal{L}_{l} + \gamma \mathcal{L}_{a},
\end{equation}
where $\alpha$, $\beta$ and $\gamma$ denote hyper-parameters, which control the importance of three losses.
It is worth noting that the alignment loss $\mathcal{L}_a$ matters in the model training. Without the $\mathcal{L}_a$, our proposed model is boiled down to learning two separated branches which is suboptimal for the fine-grained fashion similarity learning.
Besides, with the alignment loss, our proposed model, to some extent, inherits the idea of distillation~\cite{hinton2015distilling}. The global branch can be regarded as a teacher network, and the local branch plays the role of student network. Therefore, adding the alignment loss is helpful for training the local branch.
The whole two-stage training strategy is summarized in Algorithm \ref{alg_train}, where $\theta_g$ and $\theta_l$ denotes the trainable parameters of the global branch and the local branch, respectively.
It is worth mentioning that the two-stage training manner is actually easy to implement. The first training stage can be regarded as the pre-training of the global branch, which trains the global branch by minimizing the triplet ranking loss $\mathcal{L}_g$.  In the second stage, we jointly train the global and local branches by minimizing the loss defined in Eq. \ref{eq_obj}.

\algrenewcommand{\algorithmiccomment}[1]{$\triangleright$ #1}
\begin{algorithm}[t!]
\caption{Two-stage Training Strategy}
\label{alg_train}
\begin{algorithmic}[1]
\State \textbf{input:} structure of global branch $f_g$ and local branch $f_l$, triplet set $\mathcal{T}$, total training epochs $E_1,E_2$, batch size $B$, weights $\alpha,\beta,\gamma$
\State 
\State \Comment{Stage 1: line 4-10}
\For{$e\gets 1\ \textbf{to}\ E_1$}
    \For{sampled minibatch $\mathcal{B} \in \mathcal{T}$}
        \State calculate the global triplet ranking loss $\mathcal{L}_g$
        \State calculate gradients of the global branch $\nabla\mathcal{L}_g(\theta_g)$
        \State $\theta_g \gets Adam(\nabla\mathcal{L}_g(\theta_g)$)
    \EndFor
\EndFor
\State
\State \Comment{Stage 2: line 13-25}
\For{$e\gets 1\ \textbf{to}\ E_2$}
    \For{sampled minibatch $\mathcal{B} \in \mathcal{T}$}
        \State calculate the global triplet ranking loss $\mathcal{L}_g$
        \State obtain RoIs by the weakly-supervised localization
        \State calculate the local triplet ranking loss $\mathcal{L}_l$
        \State calculate the alignment loss $\mathcal{L}_a$
        \State $\mathcal{L}\gets \alpha\mathcal{L}_g + \beta\mathcal{L}_l + \gamma\mathcal{L}_a$
        \State calculate gradients of the global branch $\nabla\mathcal{L}(\theta_g)$
        \State calculate gradients of the local branch $\nabla\mathcal{L}(\theta_l)$
        \State $\theta_g \gets Adam(\nabla\mathcal{L}(\theta_g)$)
        \State $\theta_l \gets Adam(\nabla\mathcal{L}(\theta_l)$)
    \EndFor
\EndFor
\State
\State \textbf{return} trained network $f_g(\cdot)$, $f_l(\cdot)$
\end{algorithmic}
\end{algorithm}

\subsubsection{Inference} 
Once the whole network is trained, the final fine-grained fashion similarity between an image pair of $(I,I')$ \wrt an attribute $a$ is computed as the sum of their global attribute-specific similarity and local attribute-specific similarity.
Formally, the fine-grained fashion similarity \wrt an attribute $a$ is defined as:
\begin{equation}
    s(I,I'| a) = \lambda s_g(I,I' | a) + (1-\lambda) s_l(I,I', | a),
    \label{eq_sim}
\end{equation}
where $\lambda$ is a hyper-parameter to balance the importance of two branches, ranging within [0, 1]. 
Recall that our model learns multiple attribute-specific embedding spaces for fine-grained fashion similarity prediction,  the similarity \wrt a specific attribute can be easily extended to multiple attributes by summing up the similarity scores on the individual attributes.

\section{Evaluation}
To verify the viability of the proposed ASEN for fine-grained fashion similarity prediction, we evaluate it on the task of \textit{attribute-specific fashion retrieval}: Given a fashion image and a specified attribute, the task is to search for fashion images of the same attribute value with the given image.
In what follows, we firstly present the experimental setup, followed by the results of our proposed method on three datasets, \ie FashionAI \cite{zou2019fashionai}, DARN \cite{huang2015DARNdataset} and DeepFashion\cite{liu2016deepfashion}.
Ablation studies are demonstrated in Section \ref{ssec:ablation} to verify the influence of major components and hyper-parameters in the proposed model.
Then, we further visualize the obtained attribute-specific embedding spaces and attentions to investigate what has the proposed model learned in Section \ref{ssec:visual}.
Lastly, we demonstrate the potential of our proposed fine-grained similarity for general fashion reranking.

\subsection{Experimental setup}

\subsubsection{Datasets}
As there are no existing datasets for attribute-specific fashion retrieval, we reconstruct three fashion datasets with attribute annotations to fit the task, \ie FashionAI \cite{zou2019fashionai}, DARN \cite{huang2015DARNdataset} and DeepFashion~\cite{liu2016deepfashion}.

\textit{FashionAI} is a large-scale fashion dataset with hierarchical attribute annotations for fashion understanding. We choose to use the FashionAI dataset, because of its high-quality attribute annotations.
As the full FashionAI has not been publicly released, we utilize its early version released for the FashionAI Global Challenge 2018\footnote{https://tianchi.aliyun.com/markets/tianchi/FashionAI}.
The released FashionAI dataset consists of 180,335 apparel images, where each image is annotated with a fine-grained attribute. 
There are totally 8 attributes, and each attribute is associated with a list of attribute values. Take the attribute \textit{neckline design} for instance, there are 11 corresponding attribute values, such as \textit{round neckline} and \textit{v neckline}.
We randomly split images into three sets by 8:1:1, which is 144k / 18k / 18k images for training / validation / test.
Besides, for each epoch, we construct 100k triplets from the training set for model training. Concretely, for a triplet with respect to a specific attribute, we randomly sample two images of the same corresponding attribute values as the relevant pair and an image with a different attribute value as the irrelevant one. 
For validation or test set, 3600 images are randomly picked out as the query images. For each query image, the remaining images annotated with the same attribute are utilized as the candidates for retrieval. 

\textit{DARN}\cite{huang2015DARNdataset} is constructed for attribute prediction and street-to-shop image retrieval tasks. 
The dataset contains 253,983 images, each of which is annotated with 9 attributes. 
As some images' URLs have been broken, only 214,619 images are obtained for our experiments.
Images are randomly divided by 8:1:1 for training, validation and test, resulting in 171k, 43k, 43k images respectively.
Triplets are similarly sampled as that on the FashionAI dataset.
We randomly choose 100k triplets every training epoch.
For validation set and test set, images are split as query and candidate images by 1:4. 

\textit{DeepFashion}\cite{liu2016deepfashion} is a large dataset that consists of four benchmarks for various tasks in the field of clothing understanding including the category and attribute prediction, in-shop clothes retrieval, fashion landmark detection, and street-to-shop clothes retrieval. 
In this experiment, we use the category and attribute prediction benchmark.
The benchmark contains 289,222 images, 6 attributes and 1,050 attribute values, and each image is annotated with several attributes.
Similar to DARN, we randomly split the images into training, validation and test set by 8:1:1 and construct 100k triplets every training epoch. For validation set and test set, images are also split as query and candidate images by 1:4 randomly.

\subsubsection{Performance Metric}
We utilize the Mean Average Precision (MAP) and Recall@100, popular performance metrics used in many retrieval-related tasks\cite{awad2018trecvid,dong2018cross,dong2021dual,hong2017coherent,dong2021feature}. Their values are reported in percentage (\%).

\subsubsection{Implementation Details}
The proposed model is implemented using PyTorch. For the global branch, a ResNet-50 network pre-trained on ImageNet is chosen as the backbone of feature extraction. The short edges of images are first scaled to 224 while keeping the aspect ratio, then the center crop is performed before feeding to the global branch. For the local branch, we utilize a more lightweight ResNet-34 network, and the size of its input is required as 112$\times$112.
Our model is trained by the two-stage training strategy. At the first stage, the global triplet ranking loss is set with an initial learning rate of 1e-4 and a decay rate of 0.9 for every 3 epochs. The total number of epochs is 50.  At the second stage, we lower the learning rate of the global branch to 1e-5 and that of the local branch is set to 1e-4. All of them decay at a rate of 0.95 every epoch. The total number of epochs is set to be 20. For the loss function weights, we empirically set $\alpha$ to 1 and $\beta$ and $\gamma$ to 0.1. 
In Eq. \ref{eq_sim}, $\lambda$ is empirically set to 0.6.
At each epoch, we randomly sample 100k triplet examples. We train the model with a mini-batch size of 16 and optimize using Adam~\cite{kingma2014adam} optimizer.
The model that performs the best on the validation set is used for evaluation on the test set.

\subsection{Performance Comparison}

%
\begin{table*}[!t]
\renewcommand{\arraystretch}{1.3}
\caption{Performance comparison in terms of MAP on FashionAI. Our proposed ASEN model consistently outperforms the other counterparts for all attribute types.}
\label{tab_fashionAI}
\centering
\scalebox{0.9}{
\begin{tabular}{l*{9}{c}}
\toprule
\multirow{2}{*}{\textbf{Method}} & \multicolumn{8}{c}{\textbf{MAP for each attribute}} & \multirow{2}{*}{\textbf{overall MAP}} \\
\cmidrule(l){2-9}
& skirt length & sleeve length & coat length & pant length & collar design & lapel design & neckline design & neck design \\
\cmidrule(l){1-10}
Random baseline             & 17.20 & 12.50 & 13.35 & 17.45 & 22.36 & 21.63 & 11.09 & 21.19 & 15.79 \\
Triplet network             & 48.38 & 28.14 & 29.82 & 54.56 & 62.58 & 38.31 & 26.64 & 40.02 & 38.52 \\
CSN                         & 61.97 & 45.06 & 47.30 & 62.85 & 69.83 & 54.14 & 46.56 & 54.47 & 53.52 \\
$\text{ASEN}_g$     & 64.14 & 54.62 & 51.59 & 65.90 & 71.45 & 66.16 & 60.04 & 60.28 & 60.60\\
$\text{ASEN}_l$     & 50.42 & 39.93 & 40.85 & 51.87 & 67.64 & 54.38 & 50.57 & 64.11 & 50.34\\
\textit{ASEN} & \textbf{66.34} & \textbf{57.53} & \textbf{55.51} & \textbf{68.77} & \textbf{72.94} & \textbf{66.95} & \textbf{66.81} & \textbf{67.01} & \textbf{64.31}\\
\bottomrule
\end{tabular}
}
\end{table*}
\begin{table*}[!t]
\renewcommand{\arraystretch}{1.3}
\centering
\caption{Performance comparison in terms of MAP on DARN. Our proposed ASEN model still performs best.} 
\label{tab_DARN}
\centering 
\scalebox{0.8}{
\begin{tabular}{l*{10}{c}}
\toprule
\multirow{2}{*}{\textbf{Method}} & \multicolumn{9}{c}{\textbf{MAP for each attribute}} & \multirow{2}{*}{\textbf{overall MAP}} \\
\cmidrule(l){2-10}
 &clothes category&clothes button&clothes color&clothes length&clothes pattern&clothes shape & collar shape&sleeve length&sleeve shape \\
\cmidrule(l){1-11}
Random baseline                & 8.49 & 24.45 & 12.54 & 29.90 & 43.26 & 39.76 & 15.22 & 63.03 & 55.54 & 32.26 \\
Triplet network       & 23.59 & 38.07 & 16.83 & 39.77 & 49.56 & 47.00 & 23.43 & 68.49 & 56.48 & 40.14 \\
CSN                            & 34.10 & 44.32 & 47.38 & 53.68 & 54.09 & 56.32 & 31.82 & 78.05 & 58.76 & 50.86 \\
$\text{ASEN}_g$     & 38.70 & 48.91 & 52.12 & 58.44 & 54.37 & 58.50 & 36.48 & 82.42 & 59.41 & 54.30\\
$\text{ASEN}_l$     & 22.16 & 38.86 & 46.80 & 48.10 & 51.27 & 44.95 & 24.93 & 72.21 & 56.86 & 44.93\\[3pt]
\textit{ASEN}       & \textbf{40.15} & \textbf{50.42} & \textbf{53.78} & \textbf{60.38} & \textbf{57.39} & \textbf{59.88} & \textbf{37.65} & \textbf{83.91} & \textbf{60.70} & \textbf{55.94}\\
\bottomrule
\end{tabular}
}
\end{table*}
\begin{table} [tb!]
\renewcommand{\arraystretch}{1.2}
\centering
\caption{Performance comparison in terms of MAP on DeepFashion. Our proposed ASEN model still performs best.}
\label{tab_DeepFashion}
\centering 
\scalebox{0.90}{
\begin{tabular}{l*{6}{c}}
\toprule
\multirow{2}{*}{\textbf{Method}} & \multicolumn{5}{c}{\textbf{MAP for each attribute}} & \multirow{2}{*}{\textbf{overall MAP}} \\
\cmidrule(l){2-6}
 &texture&fabric&shape&part&style \\
\cmidrule(l){1-7}
Random baseline                & 6.69 & 2.69 & 3.23 & 2.55 & 1.97 & 3.38 \\
Triplet network                & 13.26 & 6.28 & 9.49 & 4.4 3 & 3.33 & 7.36 \\
CSN                            & 14.09 & 6.39 & 11.07 & 5.13 & 3.49 & 8.01 \\
$\text{ASEN}_g$     & 15.01 & 7.32 & 13.32 & 6.27 & 3.85 & 9.14\\
$\text{ASEN}_l$     & 13.66 & 6.30 & 11.54 & 5.15 & 3.48 & 8.00\\[3pt]
\textit{ASEN}     & \textbf{15.60} & \textbf{7.67} & \textbf{14.31} & \textbf{6.60} & \textbf{4.07} & \textbf{9.64}\\
\bottomrule
\end{tabular}
}
\end{table}
\begin{table} [tb!]
\renewcommand{\arraystretch}{1.2}
\centering
\caption{Performance comparison in terms of Recall@100 on FashionAI, DARN and DeepFashion.}
\label{tab_recall}
\centering 
\scalebox{0.95}{
\begin{tabular}{l*{3}{c}}
\toprule
\multirow{2}{*}{\textbf{Method}} & \multicolumn{3}{c}{\textbf{Datasets}} \\
\cmidrule(l){2-4}
 & \textbf{FashionAI} & \textbf{DARN} & \textbf{DeepFashion} \\
\cmidrule(l){1-4}
Random baseline & 5.50 & 5.27 & 2.22 \\
Triplet network & 14.29 & 7.31 & 6.68 \\
CSN & 20.21 & 14.76 & 7.92 \\
$\text{ASEN}_g$ & 24.04 & 20.06 & 9.68 \\
$\text{ASEN}_l$ & 20.46 & 14.54 & 8.17 \\
\textit{ASEN} & \textbf{25.30} & \textbf{20.74} & \textbf{10.27} \\
\bottomrule
\end{tabular}
}
\end{table}

We compare the following baselines on the three datasets: \\
$\bullet$ \textit{Random baseline}: This model sorts all candidate images randomly. As a sanity check, we include this baseline.\\
$\bullet$ \textit{Triplet network}: It learns a general embedding space to measure the fine-grained fashion similarity. It simply ignores attributes and performs mean pooling on the feature map generated by CNN. The standard triplet ranking loss is used to train the model. \\
$\bullet$ \textit{Conditional similarity network (CSN)} \cite{veit2017}:
This model first learns an overall embedding space and then employs a fixed mask to select relevant embedding dimensions with respect to the given attribute. \\
$\bullet$ $ASEN_g$~\cite{ma2020fine}: It is the conference version of our ASEN, where only the global branch is employed to learn the attribute-specific embeddings. \\
$\bullet$ $ASEN_l$: A degraded version of ASEN, where only the local branch is employed. It uses the cropped RoI with the given attribute as the input image, instead of the full image. It is worth noting that the RoI is pre-generated by our weakly-supervised localization method based on the attention result of $ASEN_g$.

Table \ref{tab_fashionAI} summarizes the MAP performance of different models on FashionAI, and the performance of each attribute type is also reported. 
Unsurprisingly, all the learning models are noticeably better than the random result. 
Among the learning based models, the triplet network which learns a general embedding space performs the worst in terms of the overall performance, scoring the overall MAP of 38.52. The result shows that a general embedding space is suboptimal for fine-grained similarity computation.
Our proposed full ASEN with a global branch and local branch performs the best. It not only demonstrates the effectiveness of our ASEN for fine-grained fashion similarity prediction, but also shows the complementary of two branches.

Additionally, using the same whole image as the input with CSN, our degraded one-branch $\text{ASEN}_g$ outperforms CSN~\cite{veit2017} with a clear margin. We attribute the better performance to the fact that $\text{ASEN}_g$ adaptively extracts feature \wrt the given attribute by two attention modules, while CSN uses a fixed mask to select relevant embedding dimensions. 
Besides, $\text{ASEN}_l$ is lightly worse than CSN in terms of the overall MAP.
The lower performance of $\text{ASEN}_l$ is likely due to that its input is cropped without any supervised bounding annotations, which may lead to wrong crops thus influence the performance. 
However, it is interesting that although $\text{ASEN}_l$ is worse in terms of the overall MAP than CSN, it achieves better in terms of certain attributes, \eg \textit{neckline design}, \textit{neck design}. Moreover, these attributes typically correspond to small regions of input images. The result shows that the local branch is better at dealing with attributes that correspond to small regions.

Table \ref{tab_DARN} and \ref{tab_DeepFashion} show the results in terms of MAP on the DARN and DeepFashion datasets.
Similarly, our proposed ASEN outperforms the other counterparts. The result again confirms the effectiveness of the proposed model for fine-grained fashion similarity prediction.
However, we find that the MAP scores on DeepFashion of all models are relatively worse compared to that on FashonAI and DARN. To some extent, we attribute it to the relatively low annotation quality of DeepFashion. For instance, only 77.8\% of images annotated with \textit{A-line} of Shape type are correctly labelled \cite{zou2019fashionai}. 
Additionally, we also summarize the performance comparison in terms of Recall@100 on FashionAI, DARN, and DeepFashion. As shown in Table \ref{tab_recall}, our proposed ASEN also performs the best.

\begin{figure*}[!t]
\centering
\includegraphics[width=0.9\textwidth]{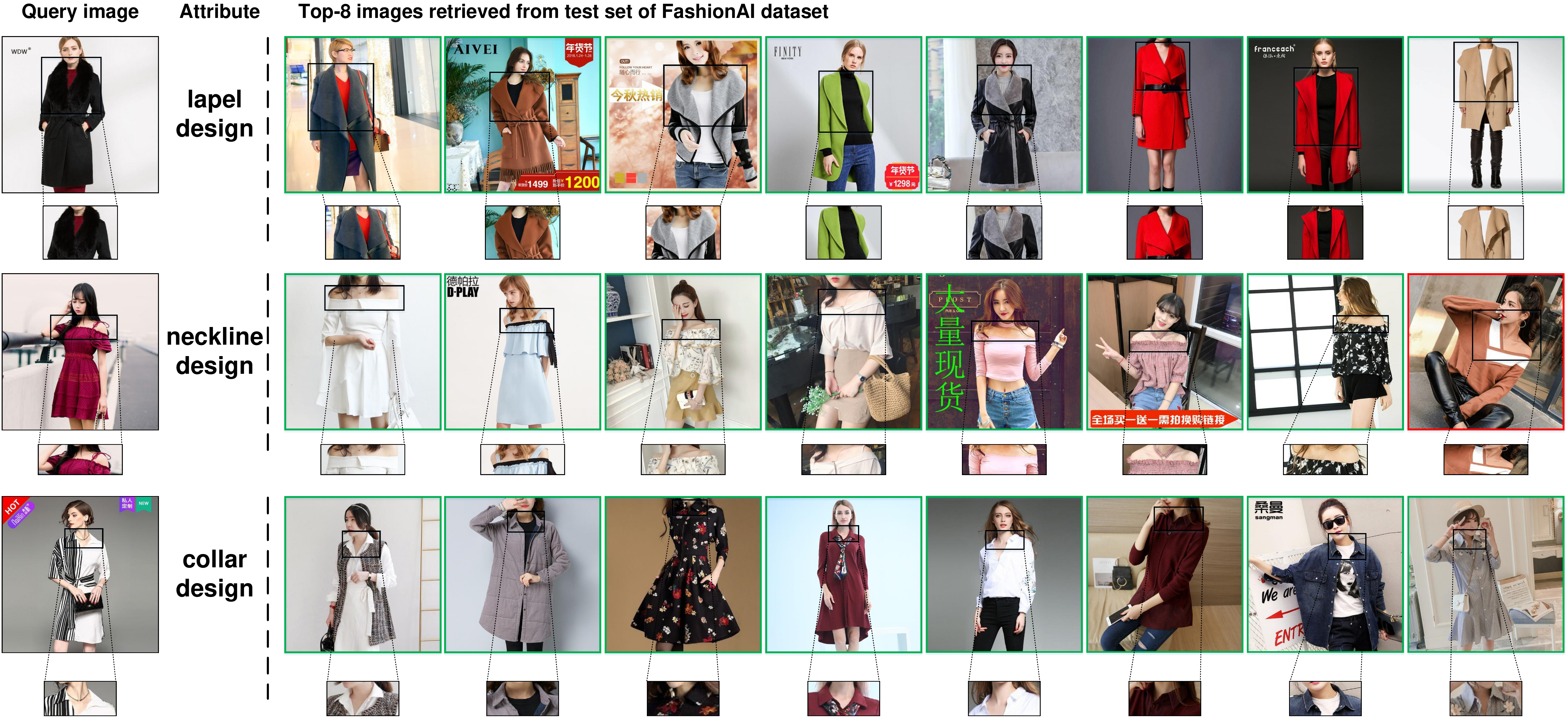}
\vspace{-3mm}
\caption{Attribute-specific fashion retrieval examples on FashionAI. Green bounding box indicates the image has the same attribute value as the given image in terms of the given attribute, while the red one indicates the different attribute values. The results demonstrate that our ASEN is good at capturing the fine-grained similarity among fashion items.}
\vspace{-3mm}
\label{fig_retrieval_examples}
\end{figure*}

%
\begin{table*}[tb!]
\renewcommand{\arraystretch}{1.3}
\caption{ Ablation studies of ASEN in terms of the model structure and the loss function on FashionAI. 
For each column, the best scores are in bold and the second best scores are underlined.
}
\label{tab_ablation}
\centering
\scalebox{0.9}{
\begin{tabular}{l*{9}{c}}
\toprule
\multirow{2}{*}{\textbf{Method}} & \multicolumn{8}{c}{\textbf{MAP for each attribute}} & \multirow{2}{*}{\textbf{overall MAP}} \\
\cmidrule(l){2-9}
& skirt length & sleeve length & coat length & pant length & collar design & lapel design & neckline design & neck design \\
\cmidrule(l){1-10}
$\text{ASEN}_g$ w/o ASA   & 62.09 & 46.18 & 49.23 & 62.79 & 67.34 & 58.07 & 46.85 & 56.20 & 54.27 \\
$\text{ASEN}_g$ w/o ACA   & 62.84 & 51.46 & 49.07 & 66.08 & 70.36 & 61.47 & 58.14 & 58.02 & 58.53 \\
$\text{ASEN}_g$           & 64.14 & 54.62 & 51.59 & 65.90 & 71.45 & 65.16 & 60.04 & 60.28 & 60.60 \\
\cmidrule(l){1-10}
ASEN w/o $\mathcal{L}_g$    & 53.73 & 13.60 & 38.55 & 57.07 & 22.59 & 22.15 & 11.44 & 21.65 & 28.82 \\
ASEN w/o $\mathcal{L}_l$    & $\underline{\text{65.63}}$ & \textbf{57.78} & $\underline{\text{54.82}}$ & $\underline{\text{68.66}}$ & 72.20 & \textbf{67.10} & $\underline{\text{66.55}}$ & \textbf{67.56} & $\underline{\text{64.08}}$ \\
ASEN w/o $\mathcal{L}_a$    & 64.95 & 55.96 & 53.76 & 67.38 & \textbf{74.12} & 66.74 & 64.51 & 66.48 & 63.05 \\
\textit{ASEN} & \textbf{66.34} & $\underline{\text{57.53}}$ & \textbf{55.51} & \textbf{68.77} & $\underline{\text{72.94}}$ & $\underline{\text{66.95}}$ & \textbf{66.81} & $\underline{\text{67.01}}$ & \textbf{64.31}\\
\bottomrule
\end{tabular}}
\end{table*}
\begin{table*}[tb!]
\renewcommand{\arraystretch}{1.3}
\caption{Ablation studies of the weakly-supervised localization method on FashionAI.}
\label{tab_localization}
\centering
\scalebox{0.9}{
\begin{tabular}{l*{9}{c}}
\toprule
\multirow{2}{*}{\textbf{Method}} & \multicolumn{8}{c}{\textbf{MAP for each attribute}} & \multirow{2}{*}{\textbf{overall MAP}} \\
\cmidrule(l){2-9}
& skirt length & sleeve length & coat length & pant length & collar design & lapel design & neckline design & neck design \\
\cmidrule(l){1-10}
\textit{ASEN} & 66.34 & 57.53 & 55.51 & 68.77 & 72.94 & 66.95 & 66.81 & 67.01 & 64.31\\
$\text{ASEN}_{\text{full}}$ & 66.51 & 55.43 & 55.37 & 67.61 & 68.58 & 62.30 & 59.09 & 57.45 & 60.81\\
$\text{ASEN}_{\text{1}}$ & 63.91 & 56.75 & 51.06 & 68.13 & 73.80 & 67.79 & 67.12 & 68.21 & 63.45\\
$\text{ASEN}_{\text{2}}$ & 65.84 & 58.04 & 54.24 & 68.74 & 72.87 & 66.94 & 66.53 & 67.31 & 64.10\\
\bottomrule
\end{tabular}}
\end{table*}

Some qualitative results of ASEN are shown in \figurename \ref{fig_retrieval_examples}. Note that the retrieved images appear to be irrelevant to the query image, as ASEN focuses on the fine-grained similarity instead of the overall similarity.
It can be observed that the majority of retrieved images share the same specified attribute with the query image. Consider the second example for instance, although the retrieved images are in various fashion category, such as \textit{dress} and \textit{blouse}, all of them are \textit{off shoulder}. 
These results allow us to conclude that our model is able to figure out fine-grained patterns in images.

\subsection{Ablation Studies}\label{ssec:ablation}
In this section, we evaluate the viability of each component in ASEN, including the attentions, loss functions, and the weakly-supervised localization. Additionally, we also investigate the influence of hyper-parameters in ASEN.

\subsubsection{Attentions}
In order to avoid the impact of the two-branch framework, we explore the attention modules on the basis of the global branch. To be specific, we respectively remove ASA and ACA from the $\text{ASEN}_g$, resulting in two degraded models, \ie $\text{ASEN}_g$ w/o ASA and $\text{ASEN}_g$ w/o ACA. The upper part of Table \ref{tab_ablation} summarizes the results. The two degraded variants obtain the overall MAP of 54.27 and 58.53, respectively. Their lower scores than $\text{ASEN}_g$ with the overall MAP of 60.60 justify the effectiveness of both ASA and ACA attentions. The result also suggests that attribute-aware spatial attention is more beneficial.

\subsubsection{Loss functions}
In this experiment, we explore the influence of the three loss functions in Eq. \ref{eq_obj}. 
Specifically, we train three ASEN variants, \ie ASEN w/o $\mathcal{L}_g$, ASEN w/o $\mathcal{L}_l$, ASEN w/o $\mathcal{L}_a$. All the variants are trained with a similar two-stage training strategy, where we keep the first stage the same as our full model but remove the corresponding loss in the second stage.
As shown in the lower part of Table~\ref{tab_ablation}, removing any loss function would result in overall MAP degeneration. It verifies the necessity of triplet ranking losses over the global branch and the local branch respectively, and the alignment loss between two branches in the second stage.
Among the three degenerated models, ASEN w/o $\mathcal{L}_g$ performs the worst, which allows us to conclude that the triplet ranking loss over the global branch is the most important among the three losses. 
Additionally, we also found that although our full model ASEN performs the best in terms of the overall MAP score, it is worse than ASEN w/o $\mathcal{L}_l$ and ASEN w/o $\mathcal{L}_a$ in several attributes.
Recall that $\mathcal{L}_l$ makes the local branch train itself by training triplets, while $\mathcal{L}_a$ makes the local branch learns from the global branches. 
We attribute this phenomenon to the similar objectives of $\mathcal{L}_l$ and $\mathcal{L}_a$, and the joint use of these two losses may limit the performance improvement and even have negative effects in some cases.

\begin{figure*}[!htb]
\centering
\subfloat[FashionAI]{\includegraphics[width=0.3\linewidth]{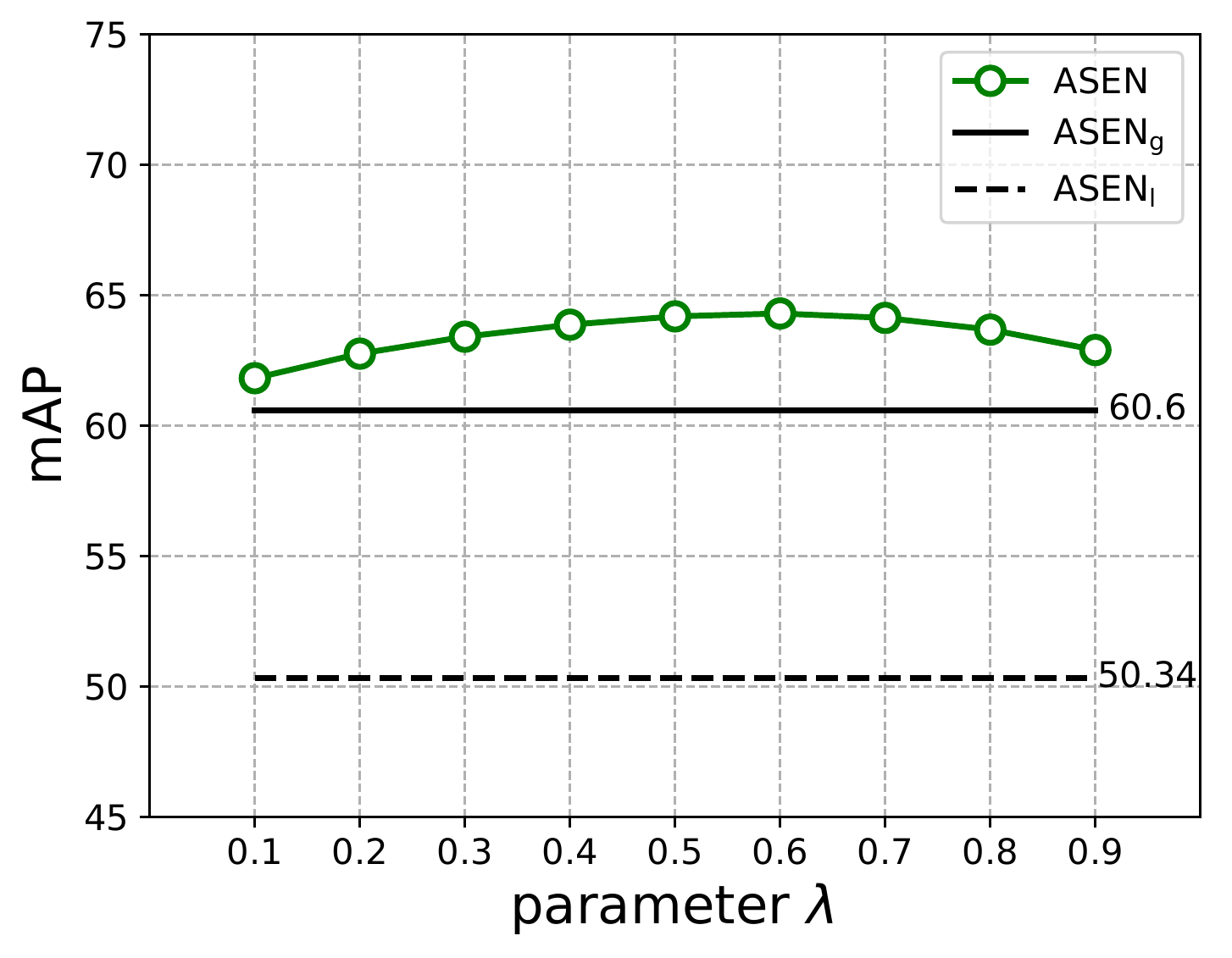}%
\label{fig_lambda_fashionai}}
\hfil
\subfloat[DARN]{\includegraphics[width=0.3\linewidth]{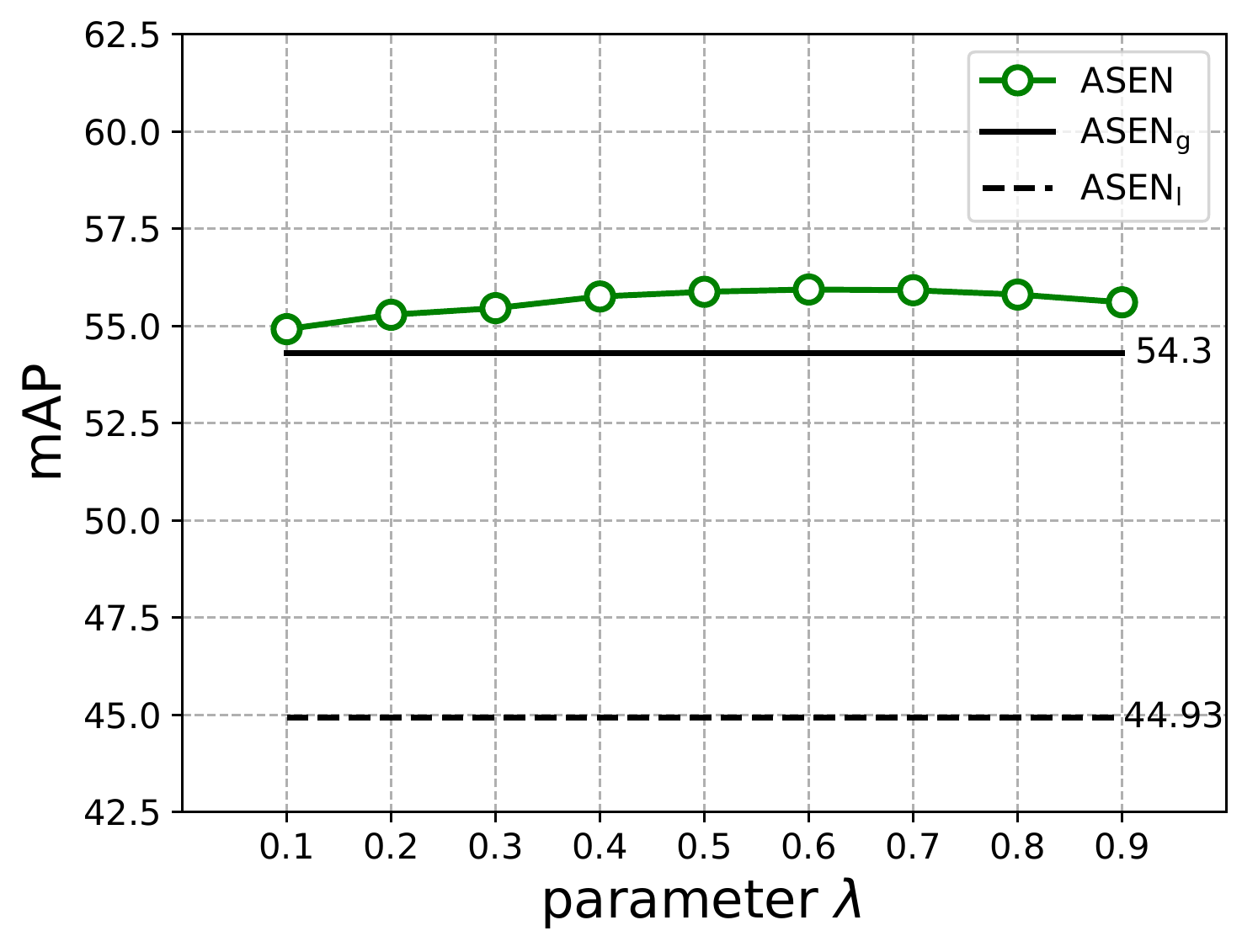}%
\label{fig_lambda_darn}}
\hfil
\subfloat[DeepFashion]{\includegraphics[width=0.3\linewidth]{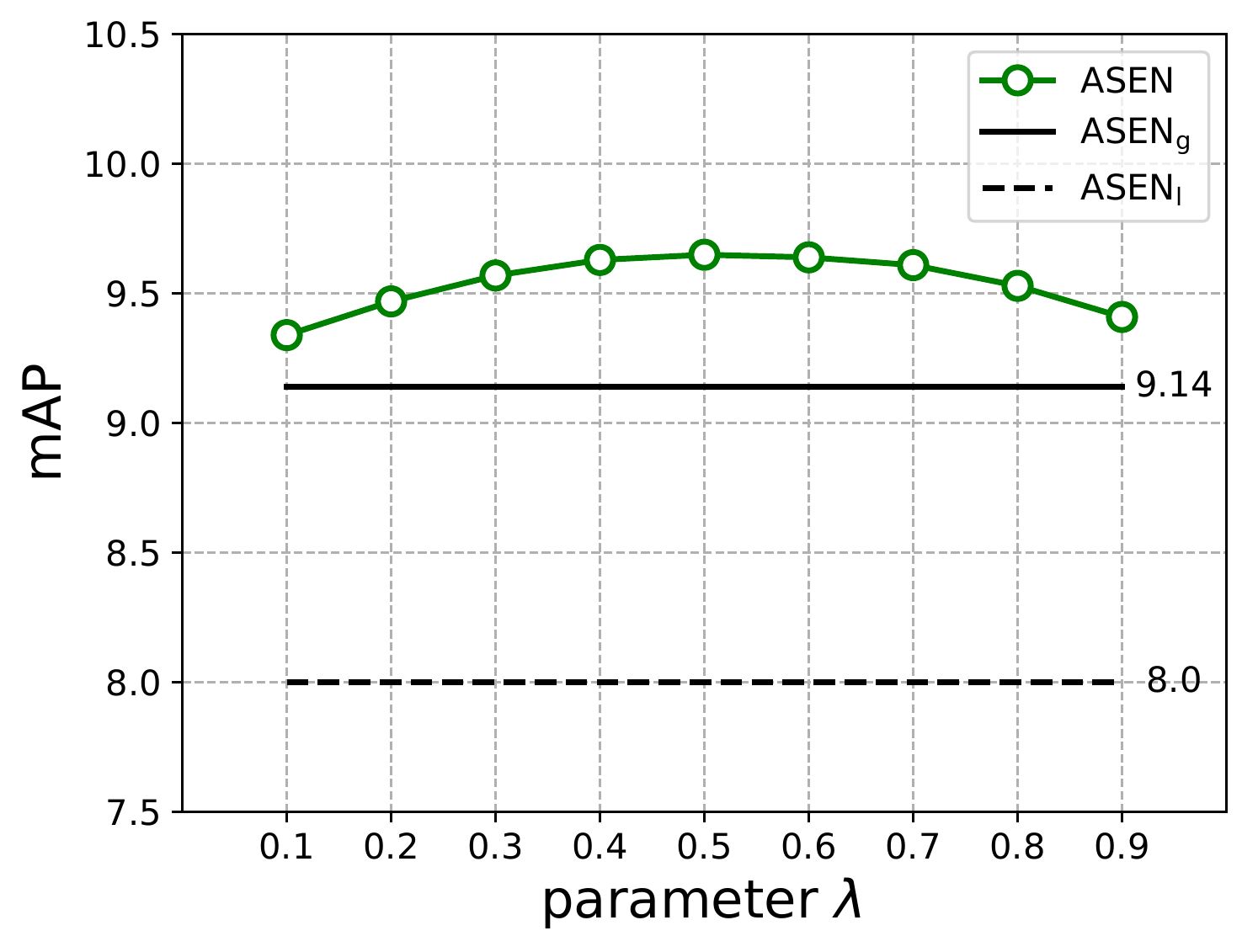}%
\label{fig_lambda_deepfashion}}
\caption{The influence of the hyper-parameter $\lambda$ of Eq. \ref{eq_sim} in ASEN on FashionAI, DARN, and DeepFashion datasets. The performance superiority of ASEN is not much sensitive to the value of $\lambda$, and the best performance is obtained when $\lambda$ is around 0.5.}
\label{fig_lambda}
\end{figure*}
\begin{figure*}[tb!]
\centering
\subfloat[coat length]{\includegraphics[width=0.4\columnwidth]{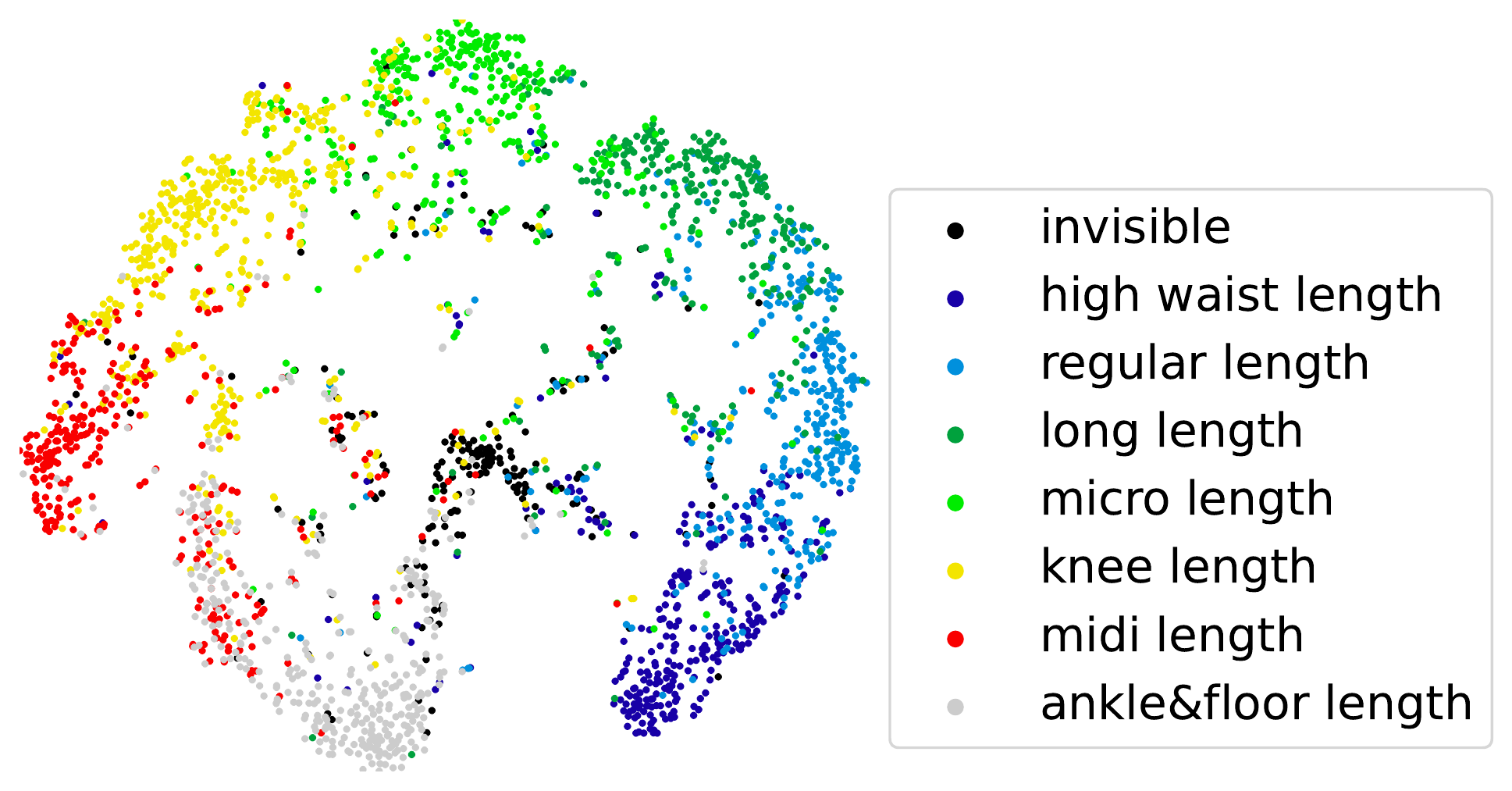}%
\label{fig_tsne_coat}}
\hfil
\subfloat[pant length]{\includegraphics[width=0.4\columnwidth]{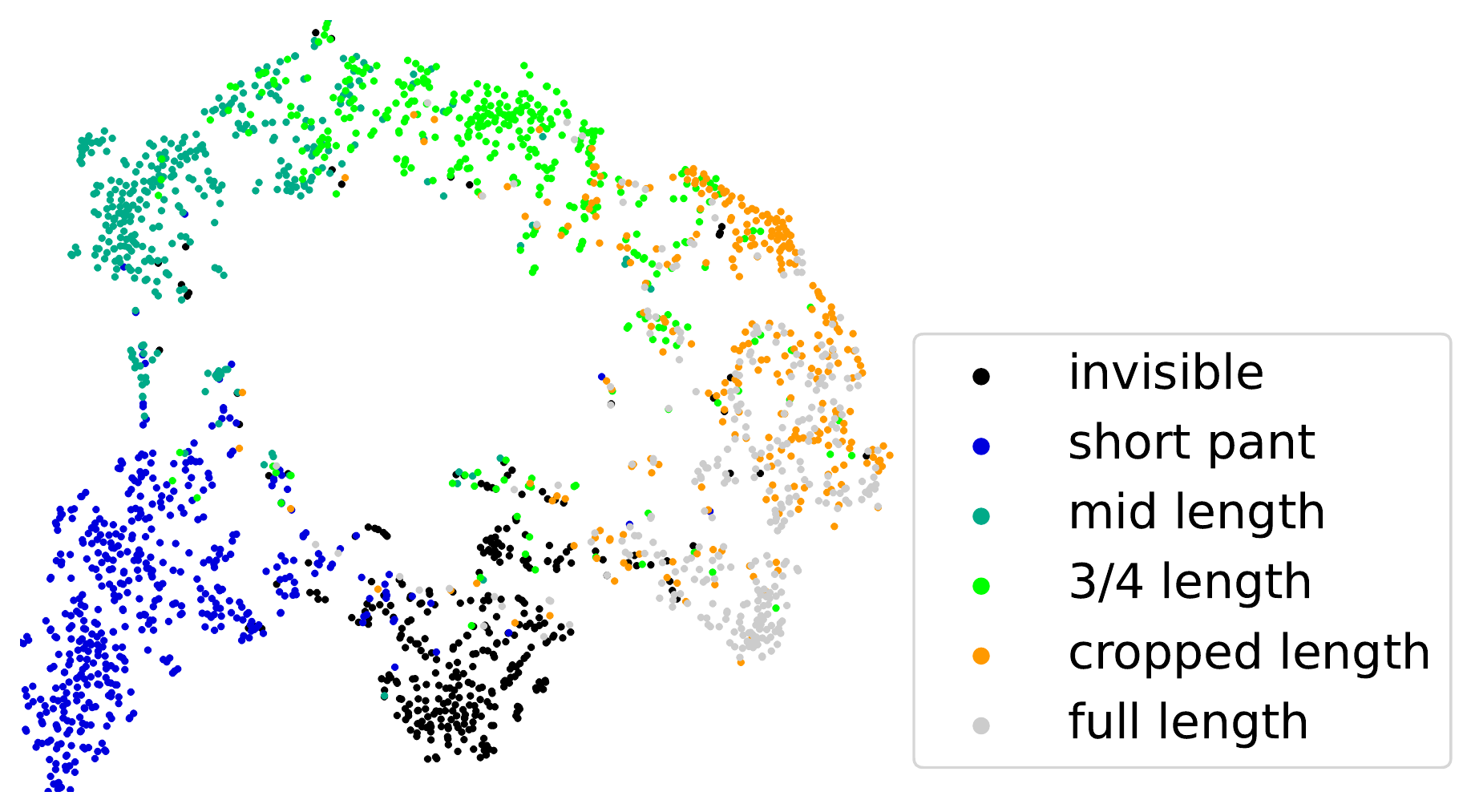}%
\label{fig_tsne_pant}}
\hfil
\subfloat[sleeve length]{\includegraphics[width=0.4\columnwidth]{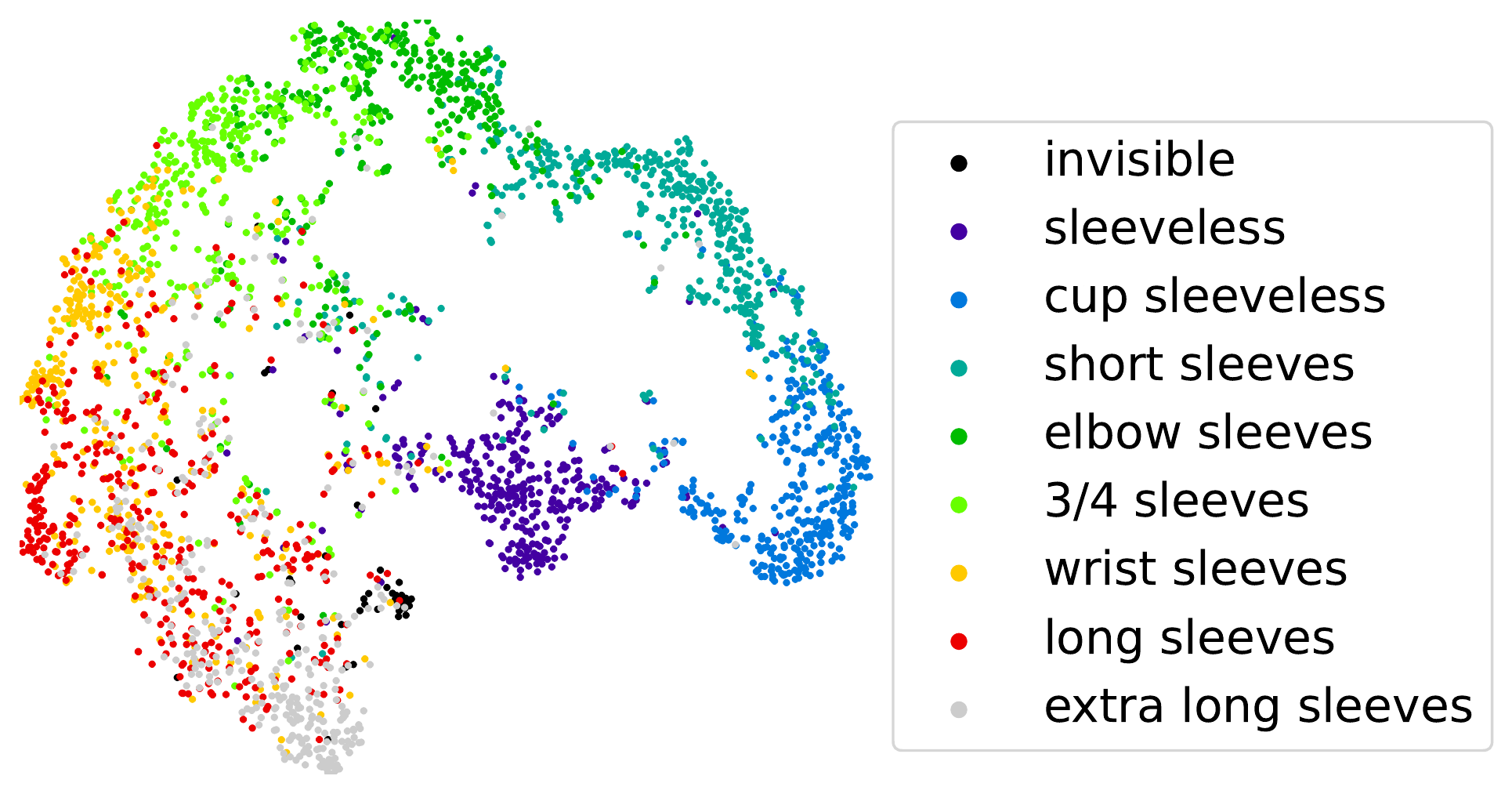}%
\label{fig_tsne_sleeve}}
\hfil
\subfloat[skirt length]{\includegraphics[width=0.4\columnwidth]{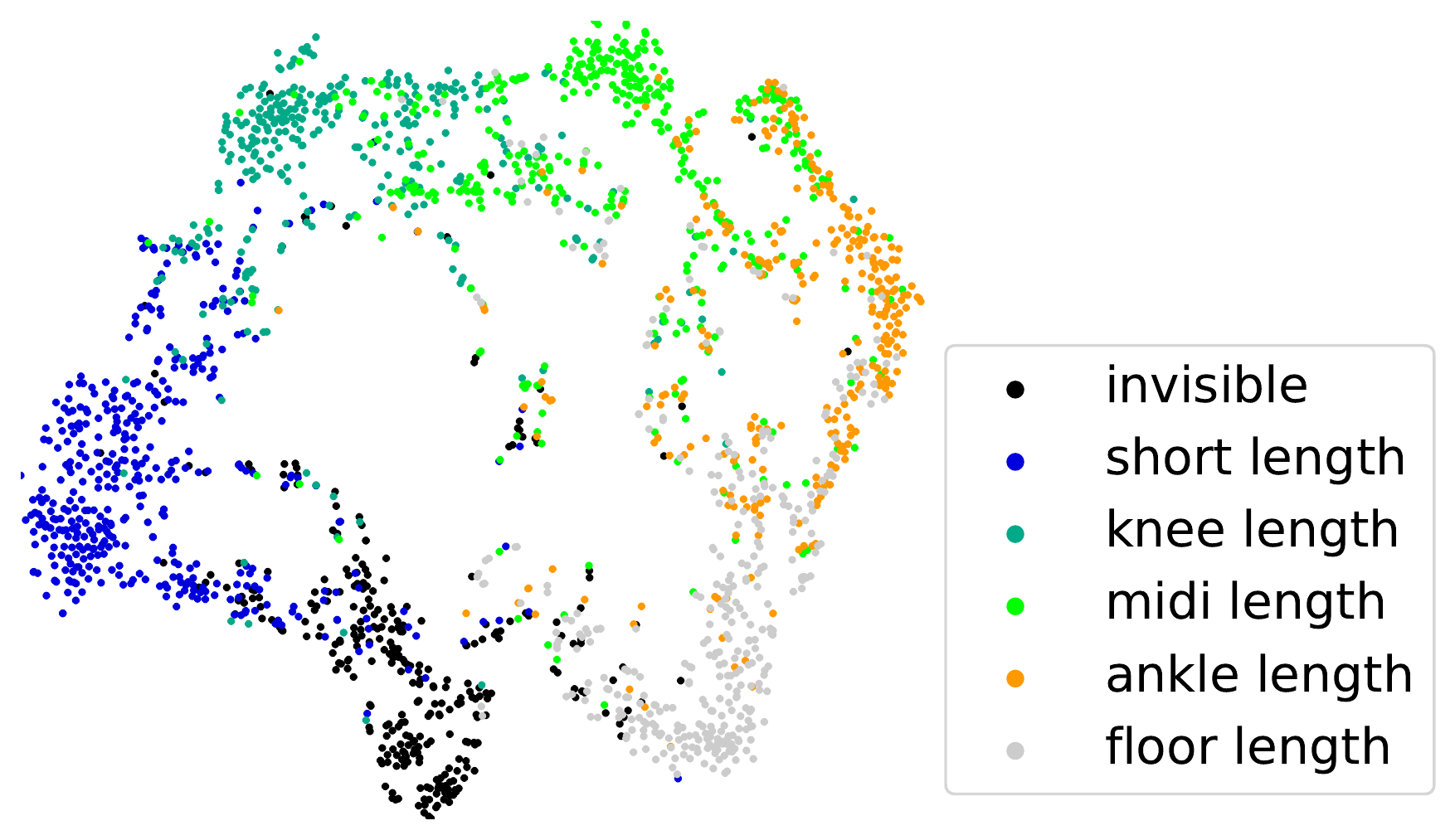}%
\label{fig_tsne_skirt}} \\
\hfil
\subfloat[lapel design]{\includegraphics[width=0.4\columnwidth]{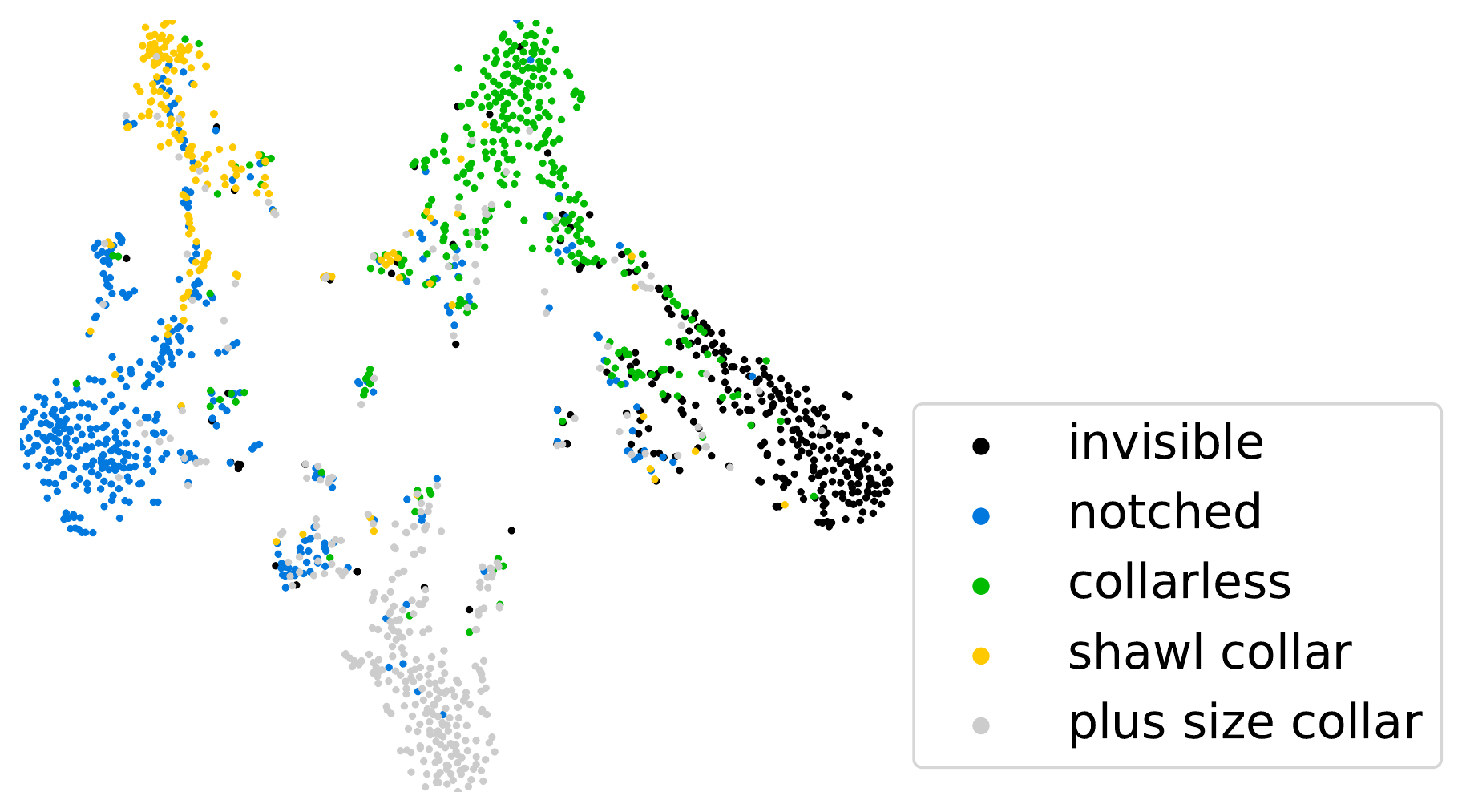}%
\label{fig_tsne_lapel}}
\hfil
\subfloat[neck design]{\includegraphics[width=0.4\columnwidth]{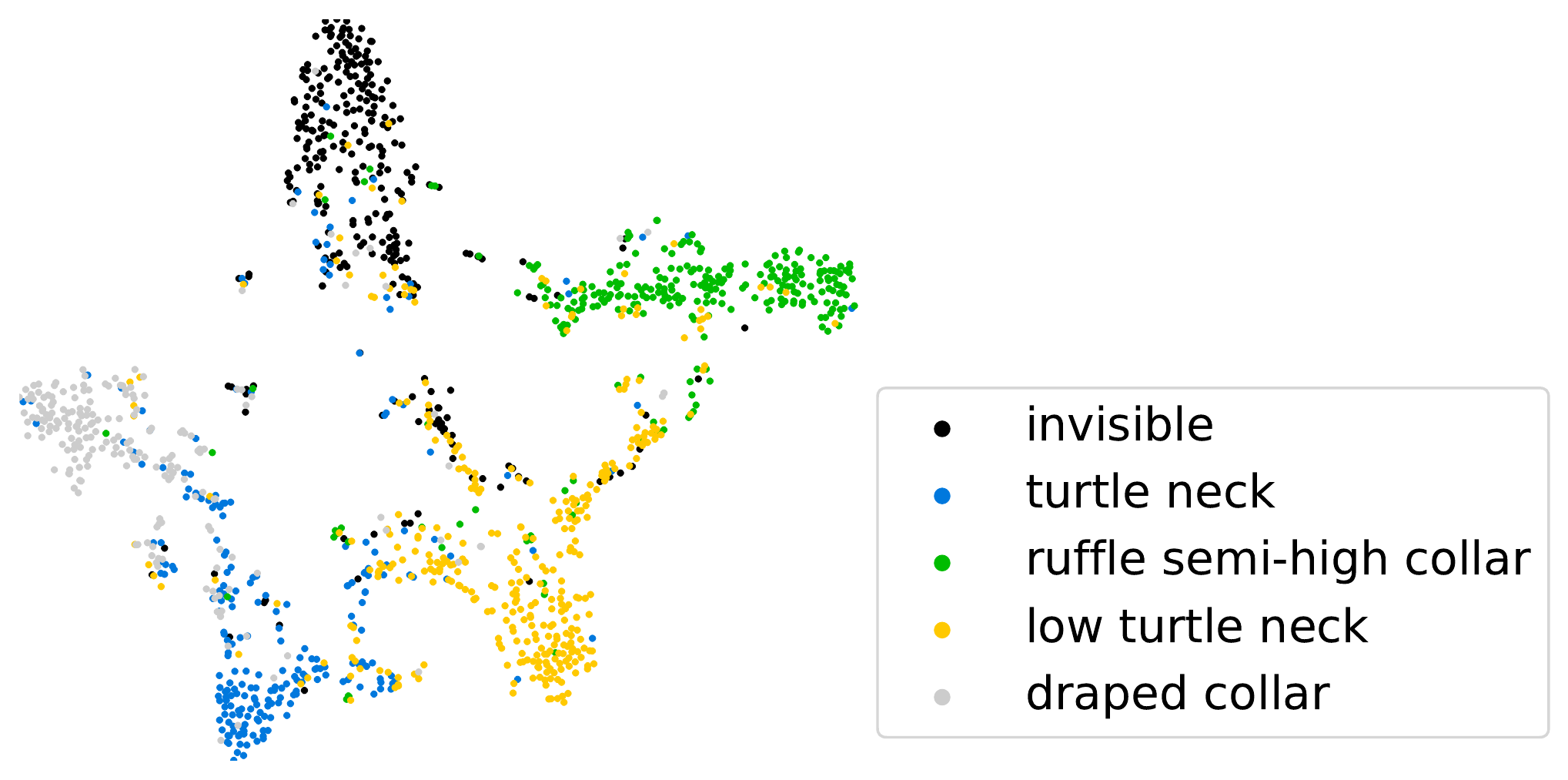}%
\label{fig_tsne_neck}}
\hfil
\subfloat[neckline design]{\includegraphics[width=0.4\columnwidth]{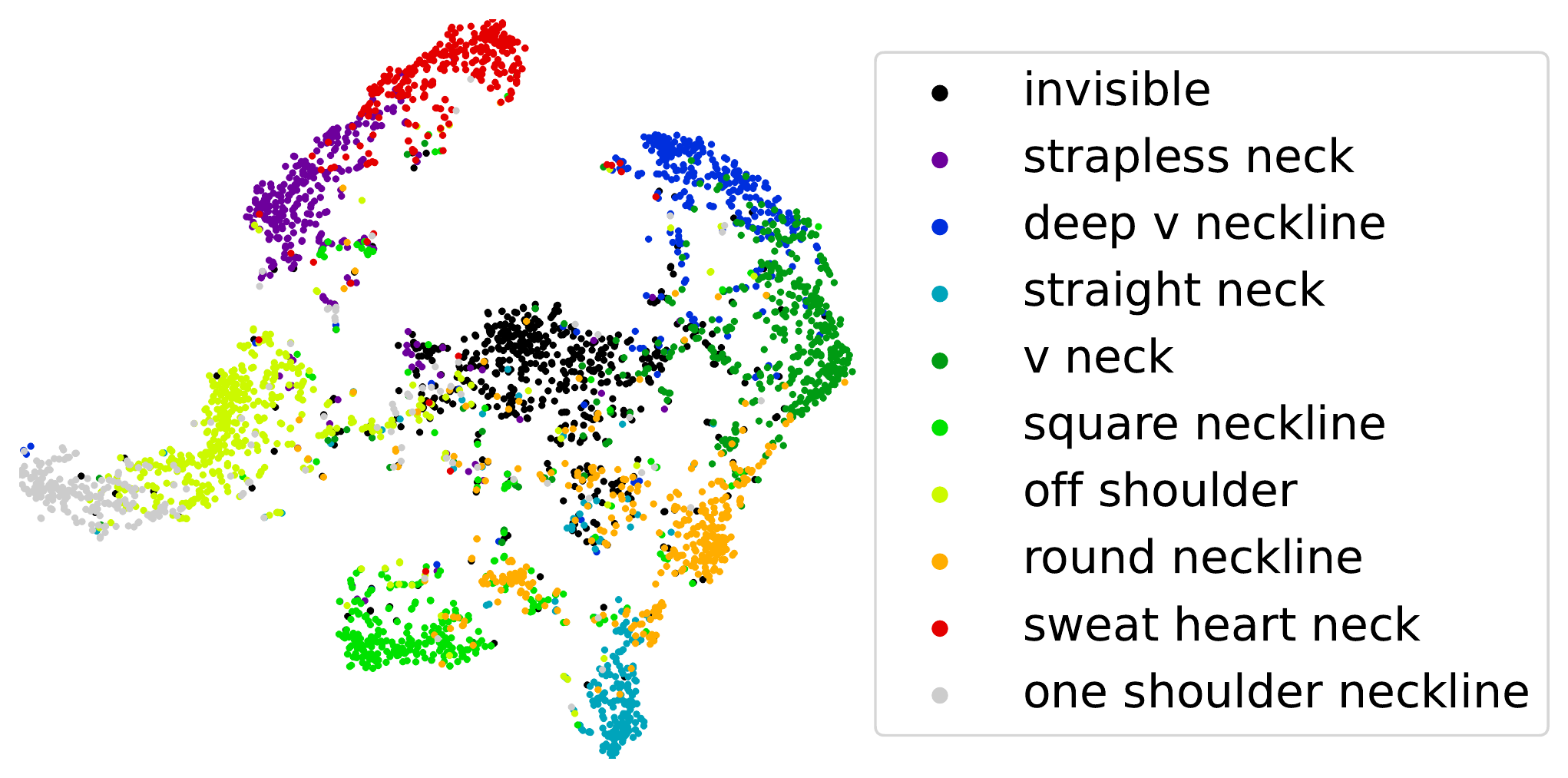}%
\label{fig_tsne_neckline}}
\hfil
\subfloat[collar design]{\includegraphics[width=0.4\columnwidth]{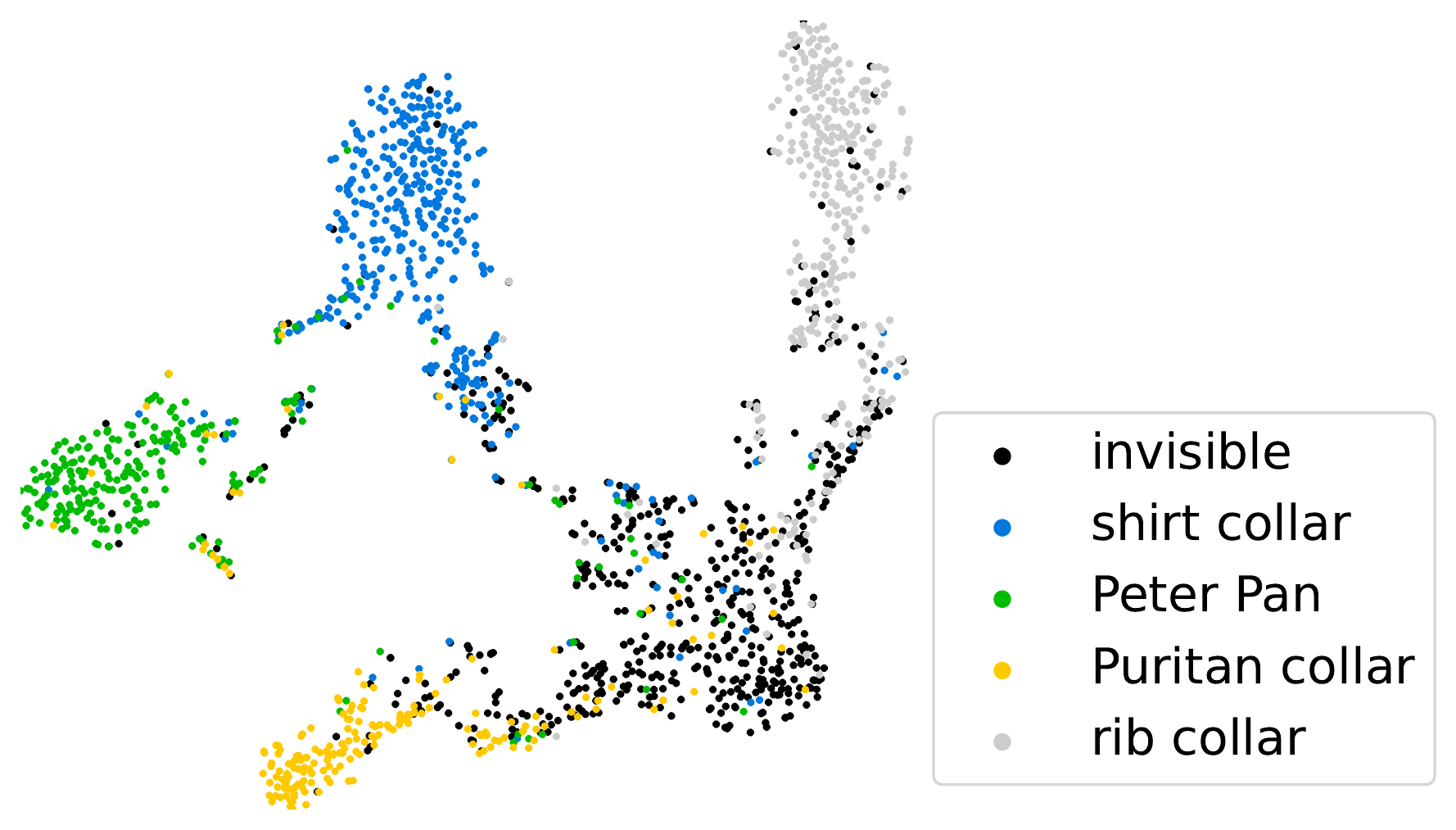}%
\label{fig_tsne_collar}}
\caption{t-SNE visualization of attribute-specific embedding spaces obtained by our proposed ASEN on FashionAI dataset. 
Dots with the same color indicate images annotated with the same attribute value. Best viewed in zoom-in.}
\label{fig_tsne}
\end{figure*}

\begin{figure}[tb!]
\centering
\includegraphics[width=0.8\columnwidth]{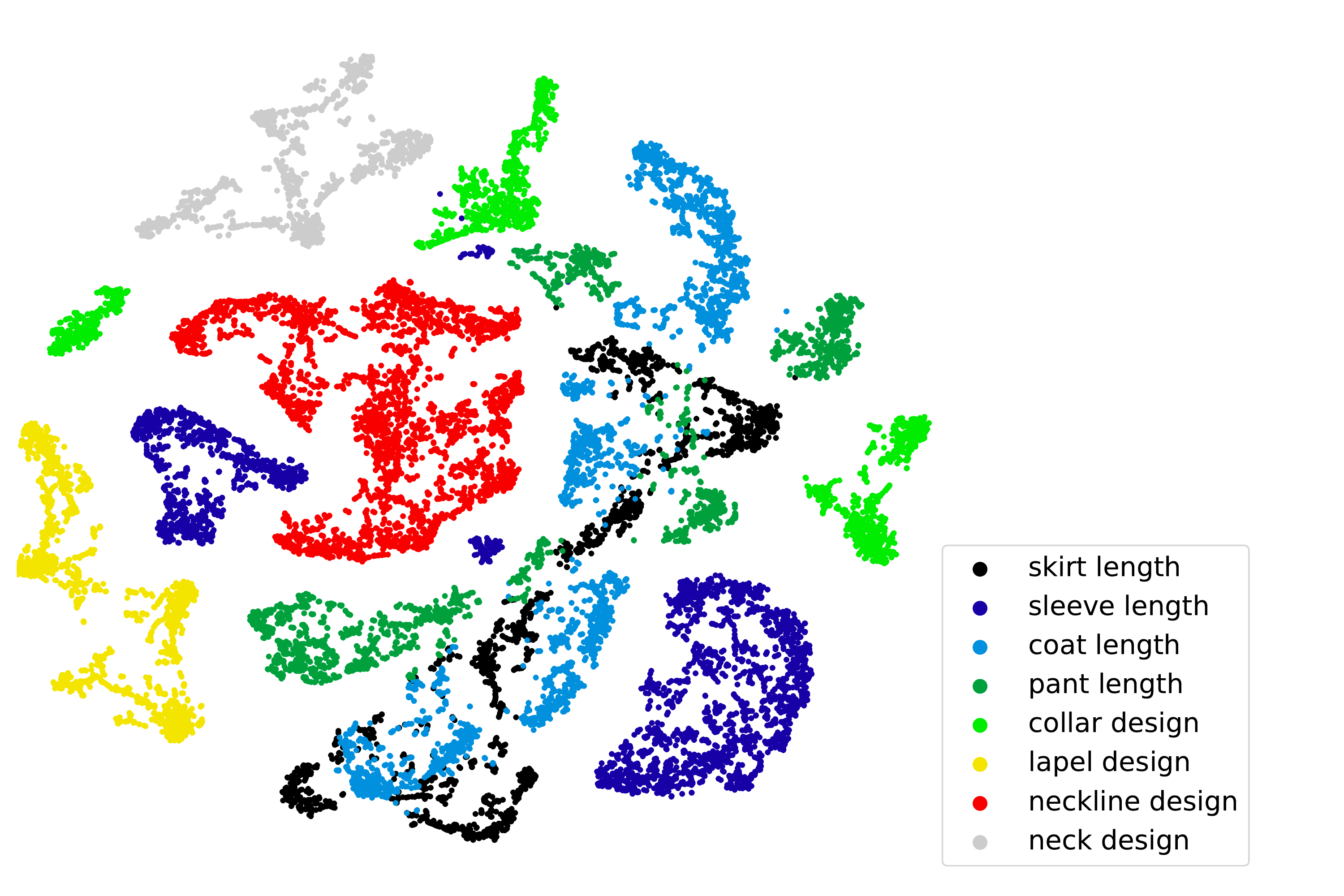}
\caption{t-SNE visualization of a whole embedding space comprised of eight attribute-specific embedding spaces learned by ASEN. Dots with the same color indicate images in the same attribute-specific embedding space.}
\vspace{-3mm}
\label{fig_tsne_attr_full_model}
\end{figure}

\subsubsection{Weakly-supervised localization}
In order to verify the effectiveness of our weakly-supervised localization for fine-grained similarity prediction, we compare it with a baseline $\text{ASEN}_{\text{full}}$ which feeds  full images into the local branch instead of RoIs. $\text{ASEN}_{\text{full}}$ is also trained with our two-stage training strategy.
As shown in Table \ref{tab_localization}, our ASEN outperforms $\text{ASEN}_{\text{full}}$ with a clear margin in terms of the overall MAP, which shows the effectiveness of our weakly-supervised localization.
For specific attribute, $\text{ASEN}_{\text{full}}$ is comparable to ASEN on four attributes about \textit{length}, as these attributes typically can be observed from the whole perspective.
For the last four attributes that tend to cover minor areas, the performance improvement of ASEN is significant. The result allows us to conclude that our weakly-supervised localization is more important for attributes that are related to minor areas.

Additionally, as previous works \cite{peng2017object,fu2017look} typically select the most discriminative region or multiple regions as local regions for fine-grained images classification, we further explore the influence of region number for our weakly-supervised localization.
Ranking regions according to their areas, we compare ASEN to $\text{ASEN}_{\text{1}}$ using the largest region and $\text{ASEN}_{\text{2}}$ using top-2 regions. As shown in Table \ref{tab_localization}, in terms of the overall MAP, our ASEN using all regions is slightly better than $\text{ASEN}_{\text{1}}$ and $\text{ASEN}_{\text{2}}$.
Interestingly, we also observe that $\text{ASEN}_{\text{1}}$ outperforms ASEN and $\text{ASEN}_{\text{2}}$ on four attributes about \textit{design}. We attribute it to that the attributes about \textit{design} typically correspond to one region of images, and utilizing more regions may introduce some noisy information. For the attributes about \textit{length} that are typically related to multiple regions, ASEN and $\text{ASEN}_{\text{2}}$ using more than one regions give better performance than $\text{ASEN}_{\text{1}}$.

\subsubsection{The influence of $\lambda$ in ASEN}
The influence of the hyper-parameter $\lambda$ in Eq. \ref{eq_sim} is studied as follows. We try $\lambda$ with its value ranging from 0.1 to 0.9 with an interval of 0.1 on three datasets. As shown in \figurename \ref{fig_lambda}, for all the choices of $\lambda$, ASEN consistently outperforms its counterparts $\text{ASEN}_g$ and $\text{ASEN}_l$. The performance of ASEN reaches its peak with an $\lambda$ of 0.6, 0.6, 0.5 on FashionAI, DARN and DeepFashion respectively. 
The results again indicate that the attribute-specific embedding learned by ASEN is stronger than that by $\text{ASEN}_g$ and $\text{ASEN}_l$, and also show that its superiority is not much sensitive to the specific value of $\lambda$.

\subsection{What has ASEN Learned?}\label{ssec:visual}

\subsubsection{t-SNE Visualization}

\begin{figure*}[!t]
\centering\includegraphics[width=0.9\textwidth]{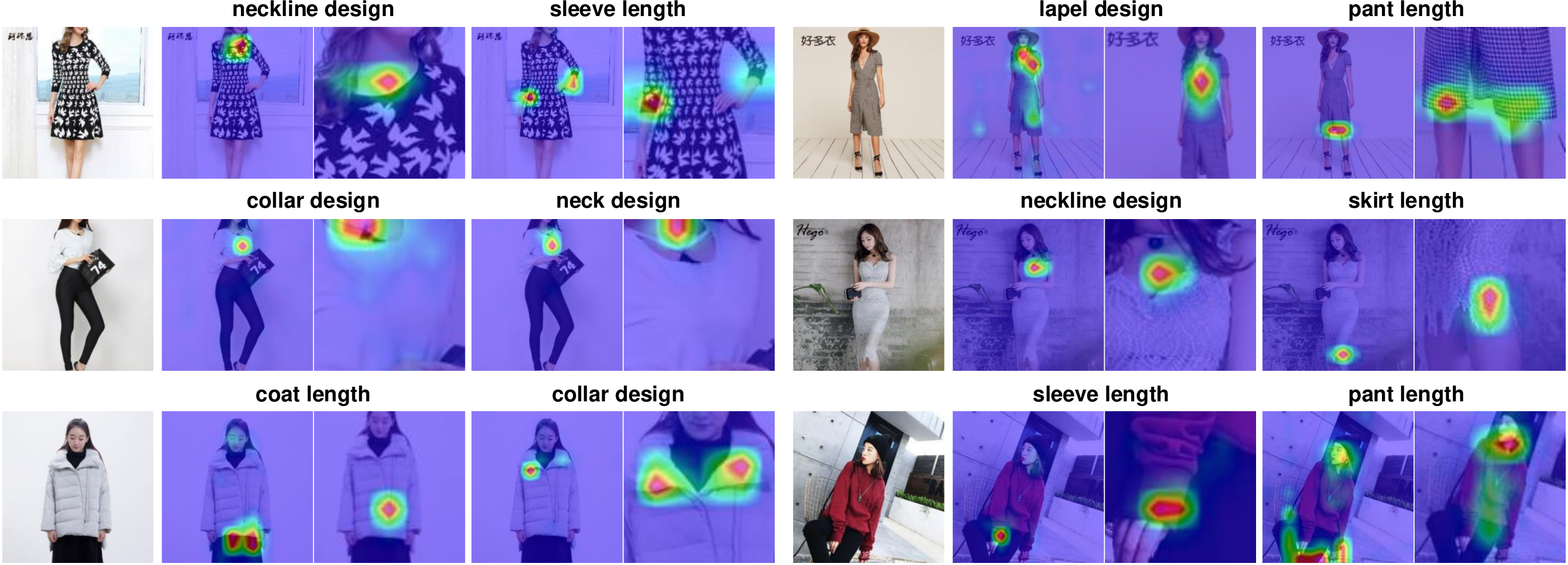}
\caption{Visualization of the attribute-aware spatial attentions of the global branch and the local branch with the guidance of a specified attribute (above the attention image) on FashionAI. For each attribute, there are two attention maps: the left one is the attention map obtained from the global branch and the right one is from the local branch .}
\label{fig_spatial_attention}
\end{figure*}

\begin{figure*}[!t]
\centering
\includegraphics[width=0.9\textwidth]{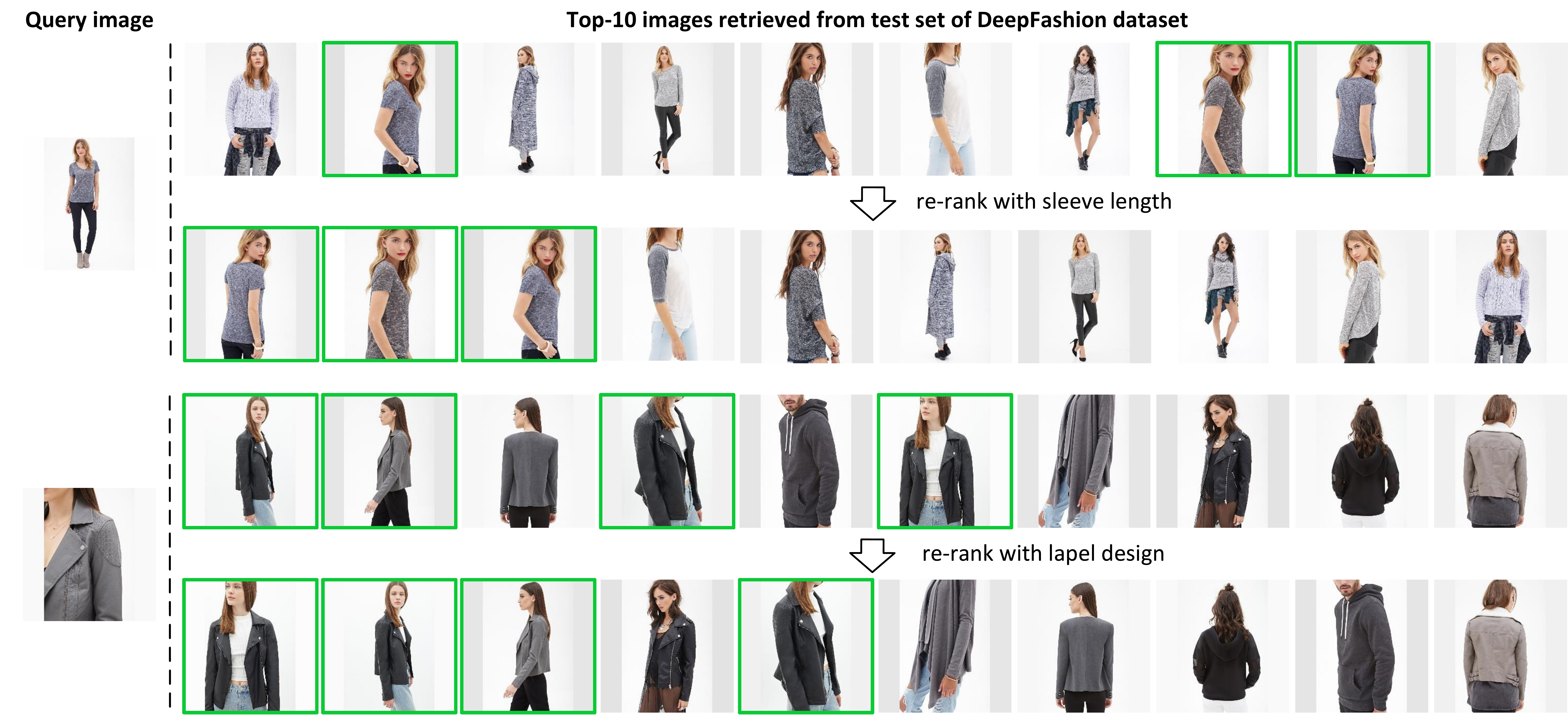}
\caption{Reranking examples for in-shop clothes retrieval on DeepFashion dataset. The ground-truth images are marked with green bounding box. After reranking by our proposed ASEN, the retrieval results become better. }
\label{fig_resort}
\end{figure*}

In order to investigate what the proposed ASEN has learned, we first visualize the obtained attribute-specific embedding spaces. Specifically, we take all test images from FashionAI, and use t-SNE \cite{maaten2008visualizing} to visualize their distribution in 2-dimensional spaces.
Fig. \ref{fig_tsne} presents eight attribute-specific embedding spaces \wrt \textit{coat}, \textit{pant}, \textit{sleeve}, \textit{skirt length} and \textit{lapel}, \textit{neck}, \textit{neckline}, \textit{collar design} respectively. Generally, dots with different colors are well separated and dots with the same color are more clustered in the particular embedding space. In other words, images with the same attribute value are close while images with different attribute value are far away. The result shows the good discriminatory ability of the learned attribute-specific embeddings by ASEN.
However, the learned embedding spaces are not perfect.
There are still some dots overlapped with that of different colors, especially for the attribute value of \textit{invisible} (black dots) that indicates the corresponding attribute does not appear or is occluded in the specific image. It may be caused by the large variety of images annotated with \textit{invisible} thus make it hard to learn.
For embedding spaces, we observe that it appears to be more overlap between different attribute values in the embedding spaces of \textit{neckline design} and \textit{collar design} than others. We attribute it to the fact that there are more images annotated with \textit{invisible} in these two embedding spaces, where over 500 images are annotated with \textit{invisible} in the embedding spaces of \textit{neckline design} and \textit{collar design} while less than 400 in other spaces.

One may ask what the relationship between the attribute-specific embedding spaces is.
To answer this question, we visualize eight attribute-specific embeddings into a whole 2-dimensional space.
As shown in Fig. \ref{fig_tsne_attr_full_model}, different attribute-specific embeddings learned by ASEN are well separated.
The result is consistent with the fact that different attributes reflect different characteristics of fashion items.

\subsubsection{Attention Visualization}
To gain further insights of our proposed network, we visualize the learned attribute-aware spatial attention maps of the global branch and the local branch. As shown in Fig. \ref{fig_spatial_attention}, the learned attention map gives relatively high responses on the relevant regions while low responses on irrelevant regions with the given attribute, showing the attention is able to figure out which regions are more important for the given attribute.
Among the global attention maps and the local attention maps, the latter typically attends on relevant areas more accurately. However, the local attention maps depend on the global attention maps. If the global attention fails to locate relevant regions \wrt the given attribute, the local attention usually also fails.
For example, in the last example with respect to \textit{pant length} attribute, the given attribute \textit{pant length} can not be reflected in the image thus the attention responses widely spread over the whole image, resulting in the wrong RoI for the local branch. Hence, the local attention map also fails to focus on the correct regions.

\subsection{The Potential for Fashion Reranking}
\subsubsection{Dataset}
In this experiment, we explore the potential of ASEN for fashion reranking.
Specifically, we consider the in-shop clothes retrieval task, in which given a query of in-shop clothes, the task is asked to retrieve the same items. We choose the in-shop clothes retrieval benchmark from the DeepFashion dataset. The benchmark has a total of 52,712 images, consisting of 7,982 different items. Each item corresponds to multiple images by changing the color or angle of shot. The images are officially divided into training, query, and gallery set, with 25k, 14k and 12k images. We keep the training set unchanged in the experiment, and respectively divide the gallery and the query images to validation and test by 1:1.

\subsubsection{Results}
Triplet network is used as the baseline to obtain the initial retrieval result. The initial top 10 images are reranked in descending order by the fine-grained fashion similarity obtained by ASEN.
We train the triplet network on the official training set of DeepFashion, and directly use ASEN previously trained on FashionAI for the attribute-specific fashion retrieval task. Fig. \ref{fig_resort} presents two reranking examples. 
For the first example, by reranking in terms of the fine-grained similarity of \textit{sleeve length}, images that have the same short sleeves with the query image are ranked higher, while the others with mid or long sleeves are ranked later.
Obviously, after reranking, the retrieval results become better. The result shows the potential of our proposed ASEN for fashion reranking.

\subsection{Complexity Analysis}
In this section, we conduct the complexity analysis of our proposed model ASEN in terms of model size and computation overhead at the inference stage.
Specifically, we measure the number of its trainable parameters and the FLOPs it takes to encode a given image via an open-source toolbox\footnote{https://github.com/sovrasov/flops-counter.pytorch}.
The total number of parameters of ASEN is roughly 21M, and its FLOPs is about 4.17G.
Additionally, we also measure the inference time of ASEN. It takes approximately 7 ms to extract the attribute-specific feature for one image on average. The inference speed is adequate for instant response. The performance is tested on a normal computer with 220G memory and a GTX 2080TI GPU.

\subsection{Limitation and Possible Improvement}
Our proposed model works with the assumption that images are associated with a specific attribute.
In the fashion retrieval scenario, especially targeting fashion copyright protection, this assumption is typically valid. For example, if someone would like to search for clothes with a certain collar design, he/she could provide an example image as the query and specify the collar design as the corresponding attribute.
However, if the assumption is invalid in some cases, our proposed model will fail to predict fine-grained fashion similarity. One possible remedy would be to discover attributes of images from their associated textual descriptions. 
Compared with the attributes, textual descriptions of fashion images are relatively easy to obtain from online shopping websites.
The viability of this remedy can be verified by the fact that attributes of DARN~\cite{huang2015DARNdataset} and DeepFashion~\cite{liu2016deepfashion} datasets are obtained by parsing textual descriptions of fashion images.
Additionally, automatically discovering the most discriminative attributes from text has also been explored for fine-grained image classification in~\cite{he2017fine,ma2019who,he2019fine}, and showing the possibility of attribute discovery from text.
In the further, we would like to improve our proposed model by discovering attributes from text.

\section{Conclusion}
In this paper, we studied the fine-grained similarity in the fashion scenario.
We contributed an \textit{Attribute-Specific Embedding Network (ASEN)} consisting of a global branch and a local branch. Both branches extracted the attribute-specific features from different perspectives, and were complementary to each other.
With two attention modules, \ie ASA and ACA, ASEN jointly learned multiple attribute-specific embeddings, and the fine-grained similarity \wrt an attribute can be computed in the corresponding space. ASEN was conceptually simple, practically effective.
Extensive experiments on various datasets support the following conclusions. 
For fine-grained similarity computation, learning multiple attribute-specific embedding spaces is better than learning a single general embedding space.
ASEN with only one global branch is more beneficial when compared with its counterpart with only one local branch. For state-of-the-art performance, we recommend two-branch ASEN with both attention modules.  The fine-grained fashion similarity is complementary to the overall similarity.

\section*{Acknowledgment}
This work was partly supported by the National Natural Science Foundation of China under No. 61772466, No. 61902347, the Zhejiang Provincial Natural Science Foundation under No. LR19F020003, No. LQ19F020002, the Zhejiang Provincial Public Welfare Technology Research Project under No. LGF21F020010, and Alibaba-Zhejiang University Joint Research Institute of Frontier Technologies.

\ifCLASSOPTIONcaptionsoff
  \newpage
\fi

\bibliographystyle{IEEEtran}
\bibliography{IEEEabrv,ref.bib}

\end{document}